\documentclass[]{interact}

\usepackage{epstopdf}% To incorporate .eps illustrations using PDFLaTeX, etc.
 \usepackage[caption=false]{subfig}% Support for small, `sub' figures and tables

\usepackage{graphicx}
\usepackage{float}
\usepackage{tabularx}
\usepackage{booktabs}
\usepackage{array} 
\usepackage{multirow}

\usepackage{cleveref}
\usepackage[numbers,sort&compress]{natbib}% Citation support using natbib.sty
\bibpunct[, ]{[}{]}{,}{n}{,}{,}% Citation support using natbib.sty
\renewcommand\bibfont{\fontsize{10}{12}\selectfont}% Bibliography support using natbib.sty
\makeatletter% @ becomes a letter
\def\NAT@def@citea{\def\@citea{\NAT@separator}}% Suppress spaces between citations using natbib.sty
\makeatother% @ becomes a symbol again

\usepackage[normalem]{ulem} % For \sout (strikeout)
\usepackage{xcolor}         % For colored text

\newif\ifrevision
\revisionfalse   % Set to \revisionfalse for clean final version

\ifrevision
  \newcommand{\add}[1]{\textcolor{black}{#1}} % old additions: plain black
  \newcommand{\remove}[1]{\textcolor{red}{\sout{#1}}} % removals stay red strikethrough

  \newcommand{\rev}[1]{\textcolor{blue}{#1}} % new additions highlighted in blue
\else
  \newcommand{\add}[1]{#1}
  \newcommand{\remove}[1]{}
  \newcommand{\rev}[1]{#1}
\fi

\theoremstyle{plain}% Theorem-like structures provided by amsthm.sty

\theoremstyle{definition}

\theoremstyle{remark}

\begin{document}

\articletype{Research Article}% Specify the article type or omit as appropriate

%\remove{Geolocation}

\title{Performance and Generalizability Impacts of Incorporating \rev{Location Encoders} into Deep Learning for Dynamic PM$_{2.5}$ Estimation}

\author{
\name{Morteza Karimzadeh\textsuperscript{a}\thanks{CONTACT Morteza Karimzadeh. Email: karimzadeh@colorado.edu}, Zhongying Wang\textsuperscript{a}, and James L. Crooks\textsuperscript{b}}
\affil{\textsuperscript{a}University of Colorado Boulder; \textsuperscript{b}National Jewish Health}
}

\maketitle

\begin{abstract}
Deep learning models have demonstrated success in geospatial applications, yet quantifying the role of geolocation information in enhancing model performance and geographic generalizability remains underexplored. A new generation of location encoders have emerged with the goal of capturing attributes present at any given location for downstream use in predictive modeling. Being a nascent area of research, their evaluation has remained largely limited to static tasks such as species distributions or average temperature mapping. In this paper, we discuss and quantify the impact of incorporating geolocation into deep learning for a real-world application domain that is characteristically dynamic (with fast temporal change) and spatially heterogeneous at high resolutions: estimating surface-level daily PM$_{2.5}$ levels using remotely sensed and ground-level data. We build on a recently published deep learning-based PM$_{2.5}$ estimation model that achieves state-of-the-art performance on data observed in the contiguous United States. We examine three approaches for incorporating geolocation: excluding geolocation as a baseline, using raw geographic coordinates, and leveraging pretrained location encoders. We evaluate each approach under within-region (WR) and out-of-region (OoR) evaluation scenarios. Aggregate performance metrics indicate that while naïve incorporation of raw geographic coordinates improves within-region performance by retaining the interpolative value of geographic location, it can hinder generalizability across regions. In contrast, pretrained location encoders like GeoCLIP enhance predictive performance and geographic generalizability for both WR and OoR scenarios. However, our qualitative analysis reveals artifact patterns caused by high-degree basis functions and sparse upstream samples in certain areas, and our ablation results indicate varying performance among location encoders such as SatCLIP vs. GeoCLIP. To the best of our knowledge, this is a first integration and systematic evaluation of location encoders in a complex, temporally dynamic estimation scenario. In addition to guiding better model development for air pollution estimation and location encoders, this study provides insights for effective incorporation of location into deep learning for geospatial predictive tasks. 
\end{abstract}


\section{Introduction}
\label{sec:Introduction}

Location plays a central role in spatial methods, including quantitative spatial analysis and geographic information science. The first law of geography by Waldo Tobler, while rooted in empirical evidence, formalizes this importance based on spatial dependence in rather simple terms: “Everything is related to everything else, but near things are more related than distant things” \cite{tobler1970computer}. Inferential statistical methods, such as linear regression models, which are originally designed for independent samples, need to be modified for application to spatial data.  Modifications are well studied and formalized in spatial statistics models, such as spatial lag or spatial error models \cite{anselin2013spatial}, or geostatistical interpolation methods such as Kriging \cite{matheron1963principles}. The list of methods for which location takes a central role is not short, and examples in this paragraph are just pointers to some of the more commonly-recognized methods.

The same principles of spatial dependence, connectivity, and location underlie deep learning (DL) architectures, which have revolutionized many domains in science and now in society under the umbrella brand of artificial intelligence (AI). In fact, the most successful and revolutionary architectures of deep learning are those that successfully leverage spatial context (i.e., spatial dependence) \cite{vaswani2017attention, krizhevsky2012imagenet}, or temporal context \cite{hochreiter1997long}, or both \cite{lecun2015deep}. These deep learning architectures have found significant performance advantages over \add{context-agnostic} prediction models. In \add{context-agnostic} prediction,  observations of $x_i$ at location $i$ are used to predict $y_i$ at the \textit{same} location. \add{Instead, modern deep learning models leverage observations made at spatial and or spatiotemporal context neighborhood to enhance estimations.}

While advances in deep learning excel at leveraging \textit{spatial} information, optimal ways of leveraging \textit{geographic} \textit{location information} remain underexplored. Note that several methods are proposed to \textit{transform}  geographic (or projected) coordinates into features for deep learning \cite{mai2022review}, however, regardless of (and beyond) the transformations used, the general higher-level approach in incorporating location into deep learning, which is still an active area of research, is our focus here.

It is crucial that we first differentiate between spatial and geographic. In the context of a Convolutional Neural Network (CNN), for example, the network learns to identify patterns within the image coordinates system, for instance around point $(x, y)$ in image coordinates, without developing an understanding of geographic location $s_i$ corresponding to latitude $\varphi$ and longitude $\lambda$ mapped to $(x, y)$.  However, how to \textit{best} incorporate location (i.e., latitude $\varphi$ and longitude $\lambda$) to maximize generalizability remains an open question, with little research quantifying the generalizability impacts of doing so using different methods on temporally-dynamic estimation scenarios. This is partly due to the gap between studies applying deep learning in geospatial sciences and environmental monitoring, where predictive performance is measured on training and testing sets with minimal interrogation of the \textit{black box} nature of deep learning, and the scarcity of fundamental studies \textit{quantifying the behavior} of (geo)location information within deep learning frameworks. Investigating the role of location in deep learning-based estimation is even more important given the emergence of ambitious \textit{location encoders} \cite{mai2023csp, klemmer2023satclip, vivanco2023geoclip}, which attempt to summarize all relevant information at location $s$ into an embedding vector $e_s$, suitable for ingestion by deep learning algorithms, for improved predictive performance, and more importantly, geographic generalizability.

Prior research has made significant progress in identifying strategies to incorporate spatial dependence \cite{lucas2023spatiotemporal} or geographic information into deep learning, including direct use of raw coordinates, hand-engineered spatial features, and learned location embeddings. Mai et al. \cite{mai2022review} provide a comprehensive survey of coordinate transformation methods—such as sinusoidal encodings, spatial tile embeddings, and kernel-based representations. Separately, contrastive learning frameworks such as \add{Contrastive Spatial Pre-training} (CSP)  \cite{mai2023csp}, SatCLIP \cite{klemmer2023satclip}, and GeoCLIP \cite{vivanco2023geoclip} have adapted CLIP-style \add{Contrastive Language-Image Pre-trainig} architectures \cite{radford2021learning} to align image and location representations in a shared latent space \add{(dubbed Contrastive Location-Image Pre-training)}, enabling pretrained models to distill contextual geographic knowledge. However, evaluations of these location encoders have been limited to static tasks such as species distribution or primitive tasks such as predicting long-term climate averages, where temporal dynamics and real-time variability are less critical, and real-world application is limited. On the other hand, studies such  as \cite{rolf2021generalizable, wang2023cross, wang2022causalgnn} have proposed robust spatial validation—such as checkerboard partitioning or distance-aware cross-validation—to more accurately assess geographic generalizability. Yet, the interplay between location encoding strategies and spatial evaluation methodology remains underexplored, particularly for dynamic, high-resolution prediction tasks.

In this paper, we take a step towards closing this gap by quantifying and analyzing the impact of incorporating geolocation information of locations $s_i$ with latitude $\varphi$ and longitude $\lambda$ in deep learning within a real-world temporally-dynamic application context of estimating daily surface-level PM$_{2.5}$ from remote sensing imagery. We first offer an in-depth discussion of different approaches to incorporating geolocation into deep learning, including simple featurization to leveraging the state-of-the-art location encoders. We then characterize the domain of surface-level air pollution estimation, including a brief overview and justification for selecting this application domain as an appropriate test-bed for this study. We then present a series of experiments and results from two complementary aspects of predictive (i.e., estimation) performance and geographic generalizability using Within-Region (WR) and Out-of-Region (OoR) evaluation scenarios. We present an ablation study to compare alternatives, and complement the experimental results with qualitative analysis of \add{estimations} in the contiguous United States. We characterize the findings and suggest directions for future research.

While the primary contribution of this paper is centered on methods of incorporating geolocation into deep learning, our work also contributes to the literature on surface-level air pollution estimation. 
%TODO here goes lit 
\add{
Recent advances in surface-level PM$_{2.5}$ estimation increasingly leverage deep learning architectures that capture spatiotemporal contextual observations, as reviewed in depth by a recent survey \cite{zhou2024deep}. Architectures combining convolutional and recurrent layers, such as CNN-LSTM models \cite{QinCnnLSTM}, have shown improved accuracy by jointly modeling spatial features and temporal dynamics. Graph-based models, including Graph Neural Networks (GNNs) and spatiotemporal Graph Convolutional Networks (GCNs), which represent spatial relationships as graphs, have been used primarily in the context of short-term forecasting at monitoring stations rather than full-coverage spatial surface level estimation \cite{zhao2021near, kim2023spatiotemporal}. Additionally, attention-based models  capture directional flow and long-range dependencies, further improving estimation or forecasting performance \cite{yu2023predicting, pathak2025novel}. These research efforts show a collective movement and growing emphasis on learning both spatial and temporal context to enhance the robustness and generalizability of PM$_{2.5}$ estimation frameworks. }

\add{ Unlike existing PM$_{2.5}$ estimation established products that rely on \add{context-agnostic} regression with raw geographic coordinates \cite{di2019ensemble}, treat spatial context as \textit{static} features like land use or proximity to emission sources \cite{ryan2007review}, or more recent research that leverage spatiotemporal \textit{context} \cite{zhou2024deep}}, our current study advances the field by explicitly interrogating the role of geolocation \add{features} in deep learning for dynamic, high-resolution air pollution estimation. Our prior work introduced a state-of-the-art Bi-LSTM with attention architecture that demonstrated superior performance, particularly during high-pollution events such as wildfires, by incorporating temporal inputs, including aerosol and meteorological data, and by integrating wildfire smoke density as a predictive covariate \cite{wang2025high, rolph2009description, mcnamara2004hazard}. This paper extends that study and evaluates and compares multiple geolocation integration strategies within the modeling framework. By evaluating both within-region (WR) and out-of-region (OoR) performance using \add{several} spatial partitioning schemes, we move beyond conventional random \add{or simpler spatiotemporal mix} validation schemes common in air pollution studies. This explicit attention to geographic generalizability and the evaluation of location encoder representations in a dynamic prediction task distinguishes our work from prior efforts, including Di et al. \cite{di2019ensemble}, Wei et al. \cite{wei2023first}, and even our own earlier study \cite{wang2025high}, which did not examine location feature representations or spatial generalization as central research questions.

To the best of our knowledge, this study is the first quantification of the impact of geolocation features and systematic evaluation of location encoders in a complex, temporally-dynamic estimation scenario, thereby expanding the empirical basis for how geolocation should be integrated into geospatial deep learning.

\section{Materials and Methods}

\subsection{Formalizing Spatiotemporal Context}
\add{
To formalize spatial and temporal context in this manuscript, consider the following:
}

\subsubsection{Spatial Context} 
Given a set of spatial locations $S = \{s_1, s_2, \dots, s_n\}$, where $s_i$ represents the coordinates of location $i$, the observation $x_i$ at location $s_i$ is influenced by its spatial neighborhood $N_s(s_i) = \{s_j : d(s_i, s_j) \le r, j \neq i \}$, where $d(\cdot, \cdot)$ can be a distance metric and $r$ is a threshold radius (or another definition of connectivity or neighborhood). A spatial prediction model can be formulated as:
\[
  y_i = f(x_i, \{x_j : j \in N_s(s_i)\})
\]
where $f(\cdot)$ is the prediction function learned by the model.
\rev{
Among such models, Convolutional Neural Networks (CNNs) \cite{lecun1998gradient} are notable.
CNNs operate on data arranged on a regular spatial grid, where features are extracted from each pixel in relation to its local neighborhood defined by a convolution window.
This architecture has been extensively used in geospatial and remote sensing applications, spanning tasks in both land \cite{zhu2017deep} and ocean remote sensing \cite{de2023model}.
}

\subsubsection{Temporal Context} 
\add{
For temporal sequences, consider a discrete time index $T = \{t_1, t_2, \dots, t_T\}$. The observation $x_i^t$ at location $i$ and time $t$ is influenced by its past observations $\{x_i^{t-k}, \dots, x_i^{t-1}\}$ for some lag $k$. A temporal prediction model is represented as:
}
\[
  y_i^t = f(x_i^t, x_i^{t-1}, \dots, x_i^{t-k})
\]
\add{
Recurrent Neural Networks (RNNs) \cite{rumelhart1986learning} and Long Short-term Memory networks (LSTMs) \cite{hochreiter1997long} belong to this group, and have been successfully adopted in geospatial applications for time-series forecasting, such forecasting the geographic spread of infectious diseases \cite{lucas2023spatiotemporal} or various remote sensing applications \cite{zhu2017tgrs}. 
Temporal models such as this actually allow implicit incorporation of absolute location $s_i$ (rather than relative location as in spatial context). For example, in geographic disease forecasting, leveraging a temporal model for each spatial unit $i$ inherently captures the disease dynamics at location $s_i$ to improve forecasting of $y_i^t$. If this implicit capturing of (geo)location is to be made explicit, then geolocation can also act as a predictive feature such that, $y_i^t = f(x_i^t, x_i^{t-1}, \dots, x_i^{t-k}, s_i)$, where $s_i$ encodes geographic location information such as latitude and longitude, spatial unit ID, or pre-learned embeddings (i.e., encodings), as discussed later in this paper. 
}

\subsubsection{Spatiotemporal Context}
\add{
Combining both spatial and temporal dimensions, the prediction at location $i$ and time $t$ depends on the spatiotemporal neighborhood:
}
\[
  y_i^t = f(x_i^t, \{x_j^{t-l} : j \in N_s(s_i), l = 0, 1, \dots, k\})
\]
\add{
In this family of models, ConvLSTMs \cite{shi2015convolutional} capture spatiotemporal context and have gained adoption in environmental monitoring \cite{le2020spatiotemporal}; Transformers \cite{vaswani2017attention} are being employed for remote sensing and geospatial applications \cite{kang2020geospatial}, and spatiotemporal graph neural networks have shown promise in spatial epidemiology \cite{wang2022causalgnn}, or even in remote sensing for building change detection \cite{Song02102023}.
}
\add{
The taxonomy above provides a non-exclusive overview of deep learning approaches incorporated in geospatial data science; for instance, sequence-to-sequence methods are not discussed here, but the reader can refer to \cite{sutskever2014sequence}. 
}

\add{
The intention is to highlight that the \add{current} revolutionary impact of deep learning methods is in large due to their ability to successfully capture spatiotemporal \add{\textit{context}}, and how there is potential for further enhancement if (geo)location information \add{is preserved within the spatiotemporal context}. 
}
\subsection{Geolocation and Deep Learning}
\label{sec:geoloc}

\add{
Deep learning is capable of capturing spatial and temporal \textit{context}; however, methods to incorporate explicit \textit{geographic location} remains an active area of research. In this section, we examine approaches to integrating geolocation—specifically latitude $\varphi$ and longitude $\lambda$—into deep learning for geospatial estimation tasks. For clarity, we focus on the role of location $s_i$ in augmenting a set of observed predictive variables $x_1$ through $x_m$, with the goal of improving both estimation accuracy and geographic generalizability. The methods discussed here can be extended to temporal and spatiotemporal context models, including the framework used in our experiments.
}

\subsubsection{Approach 1: No Geolocation Feature}
If no geolocation information is furnished to the model, then the model must learn a function $f : \mathbb{R}^m \to \mathbb{R}$ that maps the input observations $x_1, x_2, \dots, x_m$ directly to the target variable $y$, such that:
\[
y = f(x_1, x_2, \dots, x_m)
\]
where $f(\cdot)$ is learned through optimization during model training. One may ask what the purpose of intentionally leaving out the geolocation information might be: If the objective is to learn the global (albeit non-linear) mapping of observations to the target irrespective of geographic location, then excluding geographic location is justified, ensuring potentially better geographic generalizability to observations made in regions unseen by the model during training, because geolocation is not used as a predictive variable after all. Leaving out location forces the model to learn the complex and non-linear relationships between the predictors and the target, rather than using location as a predictive feature. 

\subsubsection{Approach 2: Geolocation as Geographic Coordinates Features}
This na\"{\i}ve approach involves directly including latitude $\varphi$ and longitude $\lambda$ as input features in the model. Given that geographic coordinates are cyclical in nature and measured on a spherical surface, a direct input of $\varphi$ and $\lambda$ might mislead the model in regions of transition. To handle this, a common practical technique is to transform coordinates using sine and cosine functions:
\[
\text{lat}_s = \sin(\varphi), \quad \text{lat}_c = \cos(\varphi), \quad \text{lon}_s = \sin(\lambda), \quad \text{lon}_c = \cos(\lambda)
\]
Using this transformation technique, coordinates close to each other geographically remain close in the transformed feature space, making it easier for the model to learn the geospatial relationship of observations and geolocation. The model then learns a function $f : \mathbb{R}^{m+4} \to \mathbb{R}$ that maps the observations $x_1, x_2, \dots, x_m$ along with $\text{lat}_s, \text{lat}_c, \text{lon}_s, \text{lon}_c$ to the target $y$:
\[
y = f(x_1, x_2, \dots, x_m, \text{lat}_s, \text{lat}_c, \text{lon}_s, \text{lon}_c)
\]

\add{While a variety of coordinate transformation strategies have been  surveyed before \cite{mai2022review}, we adopt the sinusoidal encoding as a representative example to evaluate the impact of geolocation features. In our contiguous U.S. study region, where latitude and longitude vary smoothly without discontinuity, the choice of transformation method has limited impact on generalizability.}

\add{Including $\varphi$ and $\lambda$ as input features enables deep learning models to learn complex, location-dependent mappings to the target variable $y$. While this can improve performance in training regions, it also risks overfitting to location-specific patterns. For example, the winning team of the AutoICE challenge \cite{stokholm2023autoice} reportedly \cite{chen2024mmseaice} used geolocation features to enhance performance in a test region that overlapped with training data \rev{(the test images in that competition were situated within the spatial extent of the training data).} Put differently, geographic overlap can inflate apparent generalization when location is used as a feature. }

On the other hand, incorporating $\varphi$ and $\lambda$ as input features can be  problematic if the predictive observations $x_1, x_2, \dots, x_m$ are weak predictors of the target $y$, resulting in the model to overly rely on the geographic coordinates to differentiate between observations, leading to overfitting on location-specific patterns rather than the actual underlying predictive relationship.  

Furthermore, using geographic coordinates can potentially harm model generalizability to other regions, even when observations are predictive of the target. Geolocation may act as a proxy for missing predictors or unmeasured covariates (similar to spatial statistical models), capturing region-specific influences that are not explicitly included as features. However, when the model is deployed in regions not represented in the training data, the absence of those region-specific patterns can lead to degraded performance, as the model's learned mapping relies heavily on location-based proxies that do not transfer well to unseen areas. 

To prevent overfitting to observations in a specific region, one potential solution is to use cross-validation or early stopping to balance overfitting and underfitting, a topic that has been studied by recent research for geospatial data \cite{wang2023cross}. However, these studies have not addressed the impacts of the use of geolocation as features in the model, and instead, focused on the broader issues of geographic generalization, even if only direct observations are involved.

\subsubsection{Approach 3: Geolocation as Pre-learned Location Encodings}
\label{subsec: Approach 3}

An alternative is to use \textbf{pretrained} location encoders to generate embeddings $e_s$ for each location $s$ and ingest those embeddings in the predictive task at hand (i.e., the downstream task). These embeddings are essentially \textit{d} dimensional vectors, akin to applying dimensionality reduction to all imaginable data at a given location (in an idealistic scenario) and reducing that data to the \textit{d} dimensions of the embedding vector. 

Location encoders aim to capture the attributes of any given location in a latent embedding vector. To better understand what these location encoders are, we must first quickly cover their inspiration: the CLIP (Contrastive Language-Image pretraining) framework \cite{radford2021learning}, initially developed by OpenAI for aligning embeddings of natural images and their captions. CLIP uses contrastive learning, training on large batches of image-caption paired data, where during (pre)training using a self-supervised approach, embeddings of positive pairs (image-caption pairs) are pulled closer together, while embeddings of negative pairs (non-matching images and captions) are pushed apart. Trained on vast amounts of naturally occurring image-caption pairs collected from the internet, CLIP results in embedding vectors for positive pairs that are \textit{aligned}. Put differently, if an image is fed to the image encoder of CLIP, and the corresponding caption is fed to the text encoder of the CLIP, they both result in embedding vectors that are very close to each other in terms of cosine similarity, and therefore, aligned in the high dimensional embedding space. CLIP-based image encoders have demonstrated superior performance over fully-supervised learning methods on benchmark datasets \cite{radford2021learning}.

Location encoders keep the image encoder in CLIP as the component of the network extracting features from (satellite or street-view) imagery, but replace the language (text) encoder with a \textit{location encoder}. This location encoder is a learnable function that operates on a high-dimensional representation of geographic coordinates produced by a fixed \textit{positional encoder} $\phi$. The positional encoder is often based on Fourier features (e.g., in GeoCLIP) or spherical harmonics (e.g., in SatCLIP), and its goal is to expand latitude $\varphi$ and longitude $\lambda$ into a higher dimension feature space that captures spatial patterns at different scales.

A general formulation of the positional encoder \rev{(based on a fourier expansion)} is:

\begin{align*}
\phi(\varphi, \lambda) = [& \ \sin(2^0 \pi \varphi), \cos(2^0 \pi \varphi), \dots, \sin(2^{k-1} \pi \varphi), \cos(2^{k-1} \pi \varphi), \\
                         & \ \sin(2^0 \pi \lambda), \cos(2^0 \pi \lambda), \dots, \sin(2^{k-1} \pi \lambda), \cos(2^{k-1} \pi \lambda)]
\end{align*}

where $k$ is the number of frequency bands used in the expansion. The location encoder is then defined as a learnable neural network $g: \mathbb{R}^{2k} \rightarrow \mathbb{R}^d$ that transforms this fixed $\phi$ representation into the final embedding vector $e_s$:

\[
e_s = g(\phi(\varphi, \lambda))
\]

During pretraining, this embedding $e_s$ is aligned with the embedding of an image taken at location $(\varphi, \lambda)$, using a contrastive loss. In other words, the image encoder distills geographic attributes (such as terrain cover, built environment, vegetation patterns) into the location encoder, effectively storing visual and contextual cues in the location embedding. In CSP \cite{mai2023csp}, $g$ is trained using ground-level imagery focused on species distribution and natural environments; in GeoCLIP \cite{vivanco2023geoclip}, Flickr images; and in SatCLIP \cite{klemmer2023satclip}, Sentinel-2 imagery.

At inference time for downstream tasks, the pretrained location encoder receives raw latitude and longitude coordinates and returns a static embedding $e_s = g(\phi(\varphi, \lambda))$. This vector encodes location-specific attributes learned during pretraining and remains fixed during downstream model training. The downstream model can then integrate $e_s$ alongside other predictive inputs (e.g., remote sensing or meteorological data), allowing it to leverage implicit spatial context—such as land use, climate zone, or infrastructure—without manually engineering or deriving these variables.

This two-stage formulation—first mapping $(\varphi, \lambda)$ into a high-frequency basis via $\phi$, then learning $g$ to align the result with image features—is foundational to modern CLIP-style location encoders. It enables geographic generalization by decoupling geographic position from learned visual cues, and supports plug-and-play use in various downstream geospatial prediction tasks.

The use of multi-frequency trigonometric functions in $\phi(\varphi, \lambda)$ allows the model to represent spatial variation at multiple scales, a key feature for learning from geospatial data with processes at different spatial resolutions. For example, consider PM$_{2.5}$ concentrations: large-scale atmospheric transport can cause regional haze events spanning hundreds of kilometers (e.g., wildfire smoke spreading across states), while at the same time, local emissions from urban centers or industrial zones create fine-scale variability within tens of kilometers.

In the Fourier-based positional encoder, low-frequency terms (e.g., $\sin(2^0 \pi \varphi)$, $\cos(2^0 \pi \lambda)$) represent coarse-scale patterns that vary slowly over space, enabling the model to recognize broad spatial gradients like continental east-west transport. In contrast, higher-frequency terms (e.g., $\sin(2^{k-1} \pi \varphi)$) allow the model to resolve fine-scale features, such as abrupt PM$_{2.5}$ spikes near city centers or downwind of industrial corridors.

The long-term vision of location encoders is to allow downstream models to benefit from location-specific knowledge without the need to collect and include all observations for every predictive task. This, theoretically should reduce the risk of overfitting, as the model is not directly \textit{memorizing} the training data but rather, learning a generalized representation of geographic context across tasks and regions. While this vision is grand, location encoders are a recent line of research, and their performance so far is evaluated on simple tasks such as average yearly temperature or other static targets, given that location encoders generate a \textit{static} vector representing locations irrespective of season or day, or any real-time observation, for that matter. In this manuscript, we focus on leveraging location encoders and their impact on generalizability and performance for estimating a dynamic, ever-changing target: daily estimation of air pollution component, PM$_{2.5}$.

\subsection{Experiments}
\label{sec:methods}

\subsubsection{Domain Characterization: Estimating Surface-level {PM$_{2.5}$}}
Estimating surface-level air pollution, and in particular, PM$_{2.5}$ is important to quantify for public health studies. Prolonged exposure to high concentrations of PM$_{2.5}$ has been \add{associated with} respiratory and cardiovascular diseases, and premature mortality \cite{reid2021daily}. 
\add{Accurate, high-resolution spatiotemporal PM$_{2.5}$ estimates have several important uses in policy-making and public health. Estimates covering recent years are used to quantify PM$_{2.5}$ human exposure in epidemiological or clinical cohort studies evaluating the impact of air pollution on specific health outcomes, which in the U.S. feed directly into regulatory decisions under the Clean Air Act. Predictions covering future days, on the other hand, are used for public health warnings, while predictions encompassing future years are used to evaluate the effects of regulatory or climatic changes.}

In the United States, the Environmental Protection Agency (EPA) has deployed a series of carefully-calibrated Air Quality System (AQS) monitors to measure the spatially and temporally variable concentrations of PM$_{2.5}$. However, public health studies require estimation of surface-level pollution at the addresses of study cohort members, while the AQS stations may be far away and spatially sparse. On the other hand,  satellite observations can be used to estimate particulates suspended in \textit{columns} of air, and therefore, the predictive task is to use satellite observations and ancillary data to derive surface-level (rather than total column) concentrations of PM$_{2.5}$. The machine learning \textit{target} is the surface-level PM$_{2.5}$ levels measured by AQS sensors. 
\add{Satellite observations are often leveraged for this purpose. Most notably, Aerosol Optical Depth (AOD), which quantifies the extinction of solar radiation by aerosol particles integrated throughout the atmospheric column, provides valuable large-scale coverage. However, it does not directly measure surface-level concentrations and exhibits a complex, nonlinear relationship with PM$_{2.5}$ due to meteorological influences, aerosol type, and vertical distribution.}

PM$_{2.5}$ is spatially and temporally variable, making remote sensing and deep learning approaches vital for generating high-resolution, daily estimates that AQS monitoring stations alone do not provide. Integrating geolocation into deep learning models has the potential to further enhance these estimates by capturing location-specific pollution sources, meteorological influences, and landuse patterns, and therefore, we believe this estimation task to be an appropriate test bed for examining the impacts of different ways of incorporating geolocation into deep learning.

We specifically focus on the task of estimating \textit{daily} concentrations, which can change from day to day, to investigate the impact of leveraging geolocation information or static location encodings, given that at the time of this writing, no spatio\textit{temporal} location encoding has been developed yet, to our knowledge. To keep our experiments further grounded, we leverage an existing and proven model with state-of-the-art performance for estimating surface-level PM$_{2.5}$ in the continental United States \cite{wang2025high}.

\subsubsection{Model Architecture}\label{sec:model_architecture}

\begin{figure}[h]
    \centering
    \includegraphics[width=0.8\linewidth]{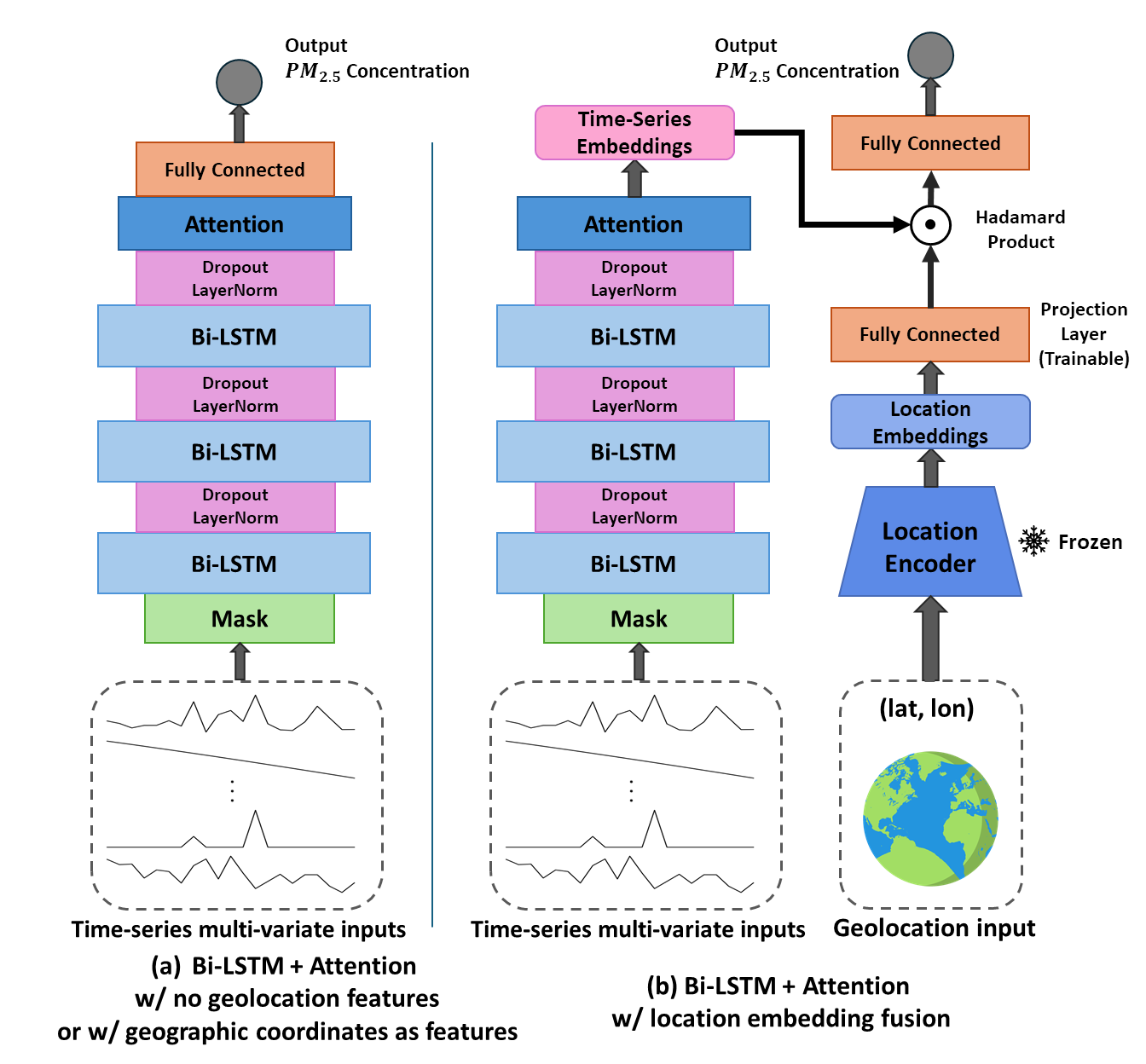}
    \caption{Model architecture for estimating surface-level PM$_{2.5}$ concentrations \add{enhanced with} a) no geolocation info, or geolocation \add{appended as geographic coordinates features}, and b) geolocation embeddings \add{vector} extracted from the pretrained location encoder \add{and fused through the Hadamard product operation}.}
    \label{fig:model_architecture}
\end{figure}

Our base model for PM$_{2.5}$ concentration prediction (Fig.\ref{fig:model_architecture}(a)) follows a Bidirectional Long Short-Term Memory (Bi-LSTM) \cite{hochreiter1997long, graves2005framewise, schuster1997bidirectional} network with Luong Attention \cite{luong2015effective}, as detailed in \cite{wang2025high}. In brief, a 21-day window of time-series multi-variate input is fed into a Bi-LSTM to capture forward and backward dependencies in PM$_{2.5}$-related features. \add{This 21-day window allows capturing of potential weeks-long persistence of PM$_{2.5}$ episodes which previous research has identified \cite{windsor2001scaling}, giving the model ample temporal context while keeping computational demands manageable. Exploratory modeling confirmed that this length achieved the lowest error, with longer windows providing no additional improvement.} 

\add{Inputs to the model include} satellite-derived Aerosol Optical Depth (AOD) \cite{lyapustin2018modis}, meteorological variables (e.g., temperature, precipitation, wind direction/speed) \cite{thornton2014daymet, abatzoglou2013development}, wildfire smoke density \cite{mcnamara2004hazard, rolph2009description}, elevation \cite{danielson2011global}, Normalized Difference Vegetation Index (NDVI) \cite{didan2015modis}, K-Nearest Neighbors Inverse Distance Weighted (IDW) PM$_{2.5}$ measurements to incorporate ground-based spatial context, as well as temporal encodings (sine and cosine of (Day of the Year, Month of the Year), and Year). 
\add{Each predictor feature was re-projected to a common 1~km grid (MODIS Sinusoidal) and, when necessary, resampled to match. Daily variables were matched by calendar date, while coarser products---such as the 16-day MODIS NDVI---were assigned to each estimation day using the nearest-available composite.}  \rev{The input datasets (predictors) and target are summarized in Table~\ref{tbl:data_summary}.}

% Column helpers
\newcolumntype{Y}{>{\raggedright\arraybackslash}X}
\newcolumntype{C}[1]{>{\centering\arraybackslash}p{#1}}

\begin{table}[h]
\caption{\rev{Datasets, features (predictors), and target used in this study.}}
\label{tbl:data_summary}
\centering
\footnotesize
\setlength{\tabcolsep}{4pt}
\renewcommand{\arraystretch}{1.2}
\begin{tabularx}{\textwidth}{
  l        % Category
  Y        % Variables / Features (flex)
  C{2.3cm} % Source (narrower)
  C{1.4cm} % Spatial
  C{1.4cm} % Temporal
  C{0.7cm} % #
}
\toprule
\textbf{Category} & \textbf{Variables / Features} & \textbf{Source (Product)} & \textbf{Spatial Res.} & \textbf{Temporal Res.} & \textbf{\#} \\
\midrule
\textbf{Target} &
Surface-level PM$_{2.5}$ concentration &
U.S. EPA AQS & Station (point) & Daily & 1 \\
\midrule
\multirow[t]{8}{*}{\textbf{Predictors}} &
Aerosol Optical Depth (Blue 0.47 $\mu$m; Green 0.55 $\mu$m) & MODIS MCD19A2.061 (MAIAC) & 1 km & Daily & 2 \\
& Meteorology: dayl, prcp, srad, tmax, tmin, vp & Daymet & 1 km & Daily & 6 \\
& Meteorology: wind direction (th), wind speed (vs) & gridMET (1/24$^{\circ}$) & $\sim$4 km & Daily & 2 \\
& Wildfire Smoke Density (WSD) & NOAA HMS Smoke & Polygon & Daily & 1 \\
& Elevation & GMTED2010 & 1 km & Static & 1 \\
& NDVI (combined) & MODIS MCD43A4 & 500 m & 16-day & 1 \\
& KNN–IDW PM$_{2.5}$ (from neighboring stations)\textsuperscript{$\dagger$} & Derived from AQS (9-NN) & N/A & Daily & 1 \\
& Temporal encodings: $\sin/\cos$ (DOY, Month), Year & Derived & N/A & Daily & 5 \\
\midrule
\multirow[t]{3}{*}{\textbf{Geolocation (variants)}} &
Latitude, Longitude & Derived (coordinates) & N/A & Static & 2 \\
& $\sin/\cos$ of lat, lon & Derived (coordinates) & N/A & Static & 4 \\
& GeoCLIP location embedding & Pretrained location encoder & N/A & Static & $512\text{-}D$ \\
\bottomrule
\end{tabularx}

\vspace{2pt}
\par\noindent\footnotesize\textit{Notes:} Reprojection/temporal matching as described in Sec~\ref{sec:model_architecture}; nearest-available NDVI composite is used for each estimation day.
\textsuperscript{$\dagger$}IDW excludes a station’s own measurement and uses only its 9 nearest neighbors to avoid data leakage.
\end{table}

These features are selected to capture the primary drivers of surface-level PM$_{2.5}$ concentrations. AOD is satellite-observed columnar aerosol loadings, which, though not a direct measure of surface-level concentrations, is predictive of surface-level pollution under certain atmospheric conditions. Meteorological variables influence pollutant dispersion, chemical transformation, and accumulation. Wildfire smoke density directly relates to episodic spikes in PM$_{2.5}$ primarily due to biomass combustion. NDVI provides information on vegetative cover, which can modulate emissions and pollutant deposition. Elevation influences local meteorology and pollutant trapping, while the KNN-IDW interpolated PM$_{2.5}$ \remove{layer} \rev{input feature} reflects observed surface-level spatial trends from monitoring stations. \rev{\rev{The supervised target in this study is the observed daily PM${2.5}$ concentration \textit{measured at AQS monitoring stations}; the KNN–IDW PM$_{2.5}$ is used only as an auxiliary predictor to convey local spatial context.}} \add{Note that the IDW-interpolated PM$_{2.5}$ excludes self-measurements of a monitoring station during training or evaluation, and instead, uses only the 9 nearest neighboring stations' ground-level measurements to calculate the interpolated values. This is to ensure there is no leakage in training data, and that estimates can be made for any point on the surface.} \rev{The interpolated feature represents only a distance-weighted neighborhood \textit{average value} and does not encode the spatial identity or exact location of any monitoring station. Because the Bi-LSTM model has no spatial receptive field and processes each sample independently, it cannot memorize or infer the true values at held-out test stations.} \rev{Also, note that because inverse-distance weighting rapidly decreases the influence of distant stations, this feature reflects primarily the immediate spatial context rather than a continuous interpolated surface. \rev{The KNN–IDW feature may be less informative in areas with sparse monitoring; however, the Bi-LSTM model uses it jointly with other predictors—such as AOD, meteorology, and vegetation—that more directly capture the pollution level in those regions.}} \remove{Temporal encodings help capture seasonal patterns in emissions.}

\add{Because AOD is a key predictor of surface PM$_{2.5}$ and a time-series of it is fed to the model for every pixel, missingness caused by cloud or snow cover were first imputed with a lightweight random forest regressor trained on the same set of predictors (excluding the KNN-IDW PM$_{2.5}$ and the target itself), as detailed in \cite{WangImpute2023}. After this step, a single masking layer flags any residual missing entries across all predictors, enabling the Bi-LSTM to downweight them during training rather than treat them as observed values \cite{che2018recurrent}.} 

Layer normalization and dropout \cite{ba2016layer} are applied after each Bi-LSTM layer to stabilize training and mitigate overfitting. The Luong attention mechanism \cite{luong2015effective} then learns to focus on the most informative timesteps within the 21-day window by computing a context vector as a weighted average of the hidden states. The representation is then passed through a fully connected layer to make the final prediction for PM$_{2.5}$. We refer the reader to \cite{wang2025high} for detailed information on data processing, implementation, long-term evaluation of the model and comparison to multiple other baselines and benchmarks. \add{What is different in this manuscript is the ways in which geolocation features are fed to the model, and the way performance is evaluated under regional conditions. Specifically, we use several configurations in regional partitioning for creating training and testing splits as detailed in the Section \ref{sec:Results}, and for each configuration, train the model with four general variations---differing in the way geolocation is incorporated:}

\begin{enumerate}
    \item \textbf{No geolocation input}: The model does not receive geolocation $(\varphi,\lambda)$ information, similar to Approach 1  described in \cref{sec:geoloc}, forcing the model to purely rely on time-series observations for estimating the target. 
    \item \textbf{Direct geolocation input}: We append the raw latitude and longitude values $(\varphi,\lambda)$  to the time-series feature vector at each timestep, similar to Approach 2  described in \cref{sec:geoloc}, allowing the model to rely on both the time series observations as well as geolocation for estimating the target. 
    \item \textbf{Sinusoidal transformation of geographic coordinates}: Also similar to Approach 2 in \cref{sec:geoloc}, but with a transformation applied first: instead of passing latitude and longitude values directly in angular units, we apply sinusoidal transformations to preserve their cyclical properties: 
    \[
\text{lat}_s = \sin(\varphi), \quad \text{lat}_c = \cos(\varphi), \quad \text{lon}_s = \sin(\lambda), \quad \text{lon}_c = \cos(\lambda)
\]
    These four transformed features are then appended to the time-series inputs, allowing the model to rely on both the time series observations as well as transformed geolocation for estimating the target. 
    \item \textbf{Fusing location encoder embeddings}: We integrate \textbf{pre-learned} location embeddings from the pretrained location encoder  GeoCLIP \cite{vivanco2023geoclip}. GeoCLIP is a location encoder trained using a CLIP-style contrastive learning framework to align geotagged Flickr images with geographic coordinates (\cref{subsec: Approach 3} above covers details of this approach). Its image encoder is a Vision Transformer (ViT-B/16), while the positional encoder is a Fourier feature mapping of latitude and longitude, capturing multi-scale spatial patterns. During pretraining by the original developers \cite{vivanco2023geoclip}, GeoCLIP was trained on over 1.2 million image-location pairs globally sampled from the YFCC100M dataset, with relatively denser coverage in urbanized and photo-rich regions, which is important for PM$_{2.5}$  estimation in areas more prone to generate emissions. We will contextualize GeoCLIP against other available location encoders in \cref{subsec:Ablation}. 
    
    As shown in Fig.\ref{fig:model_architecture}(b), the location encoder outputs a vector $e_{\text{s}}$, summarizing geographic information for a given latitude and longitude. In line with the original intent of location encoders, we freeze the location encoder remains during training to preserve the learned spatial representations. A trainable projection layer is applied to $e_{\text{s}}$ to align its dimensionality with the time-series embeddings $e_{\text{ts}}$ produced by the Bi-LSTM with Attention branch. Specifically, let $e_{\text{proj}} = \text{Projection}(e_{\text{s}})$ denote the projected location embeddings. We then fuse $e_{\text{proj}}$ and $e_{\text{ts}}$ via a Hadamard product:
\begin{align*}
    e_{\text{fused}} = e_{\text{ts}} \odot e_{\text{proj}}
\end{align*}

where $\odot$ denotes Hadamard product. This operation allows the model to learn the interactions between temporal patterns and geographic context. The fused embedding $e_{\text{fused}}$ is then passed to a final fully connected layer that regresses the surface-level PM$_{2.5}$ concentrations, allowing the model to rely on both the time-series observations as well as pre-learned  embeddings for the given location.

\end{enumerate}

\subsubsection{Model Training}
The training logistics settings remain the same as the original model described in \cite{wang2025high}. Our region, similarly, encompasses the Continental United States (CONUS), but we limit our daily estimation to all days in 2021 to keep computations manageable for this paper. Inputs are normalized feature-wise to the range of [-1,1] using a MinMax Scaler. We employ the Adam optimizer \cite{kingma2014adam} with an exponential learning rate scheduler \cite{sutskever2013importance} (initial learning rate of $1e^{-3}$, decay factor of 0.8 every 30,000 steps). All models are trained for 100 epochs with a batch size of 256, and Huber loss is used to mitigate the impact of outliers in the PM$_{2.5}$ data \cite{huber2011robust}. 

\section{Results}
\label{sec:Results}
We assess model performance and generalizability using three main evaluation settings, followed by ablation and qualitative analyses.  In all experiments, we report the coefficient of determination ($R^2$), root mean square error (RMSE), \rev{and mean bias error (MBE)} as performance metrics for average of five training runs. \rev{Complete evaluation tables are provided in Appendix Table~\ref{tab:S1_random}-\ref{tab:S4_cb16}. We provide key metrics in the following Sections.}  Please note that the results \textbf{within} each setting can be compared, not across settings, to ensure the number of training samples, spatial distributions, and all other configurations remain similar. 

\subsection{Within-Region (WR) Evaluation}
Within-region scenarios are ones where the test locations are geospatially in-distribution in comparison to training locations. In other words, these experiments are designed to see if incorporating geolocation information enhances the model's predictive performance and adds geospatial \textit{interpolative value}. 

\textbf{Random Test Set Evaluation}: In this setting, 10\% of the samples are randomly selected as the test set. Because the model uses multivariate time-series inputs, test samples may include locations whose temporal observations (at different timesteps) were seen during training.

\textbf{Spatial Test Set Evaluation}: To evaluate the model's ability to predict at unseen locations, we create a spatial test set by randomly dropping 10\% of the unique locations from the dataset. While this approach tests spatial generalizability and is common for air pollution estimation evaluation \cite{reid2021daily},  it is worth noting that the random selection process does not enforce spatial clustering—meaning that the training set may still include locations that are geographically close to test locations. Put differently, this experiment still tests for Within-Region (WG) performance. 

\begin{figure}[htbp]
    \centering
    \subfloat[Random test set metrics\label{fig:randomcv}]{
        \includegraphics[width=0.85\textwidth]{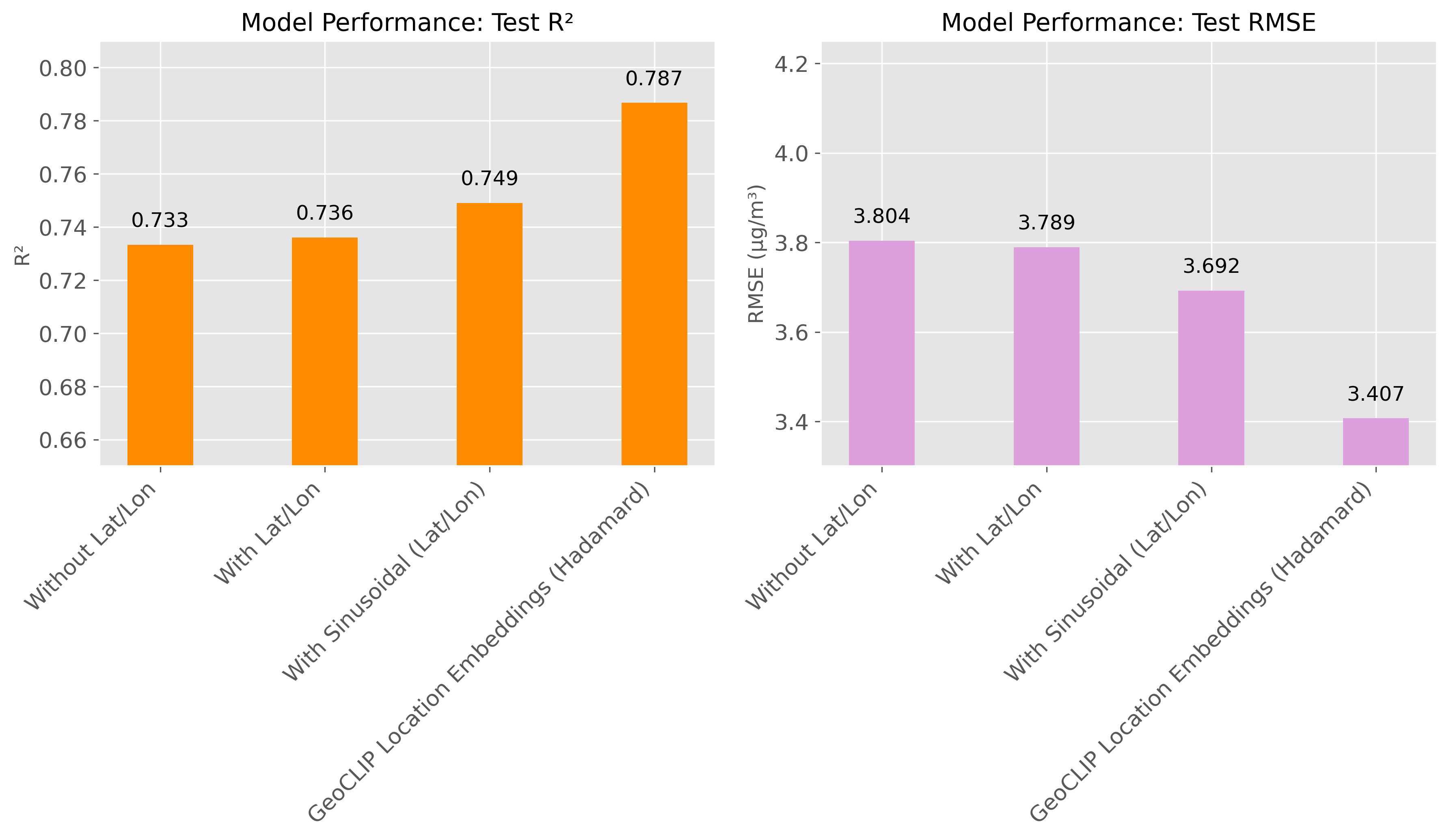}
    }\vspace{0.2cm}

    \subfloat[Spatial test set metrics\label{fig:spatialcv}]{
        \includegraphics[width=0.85\textwidth]{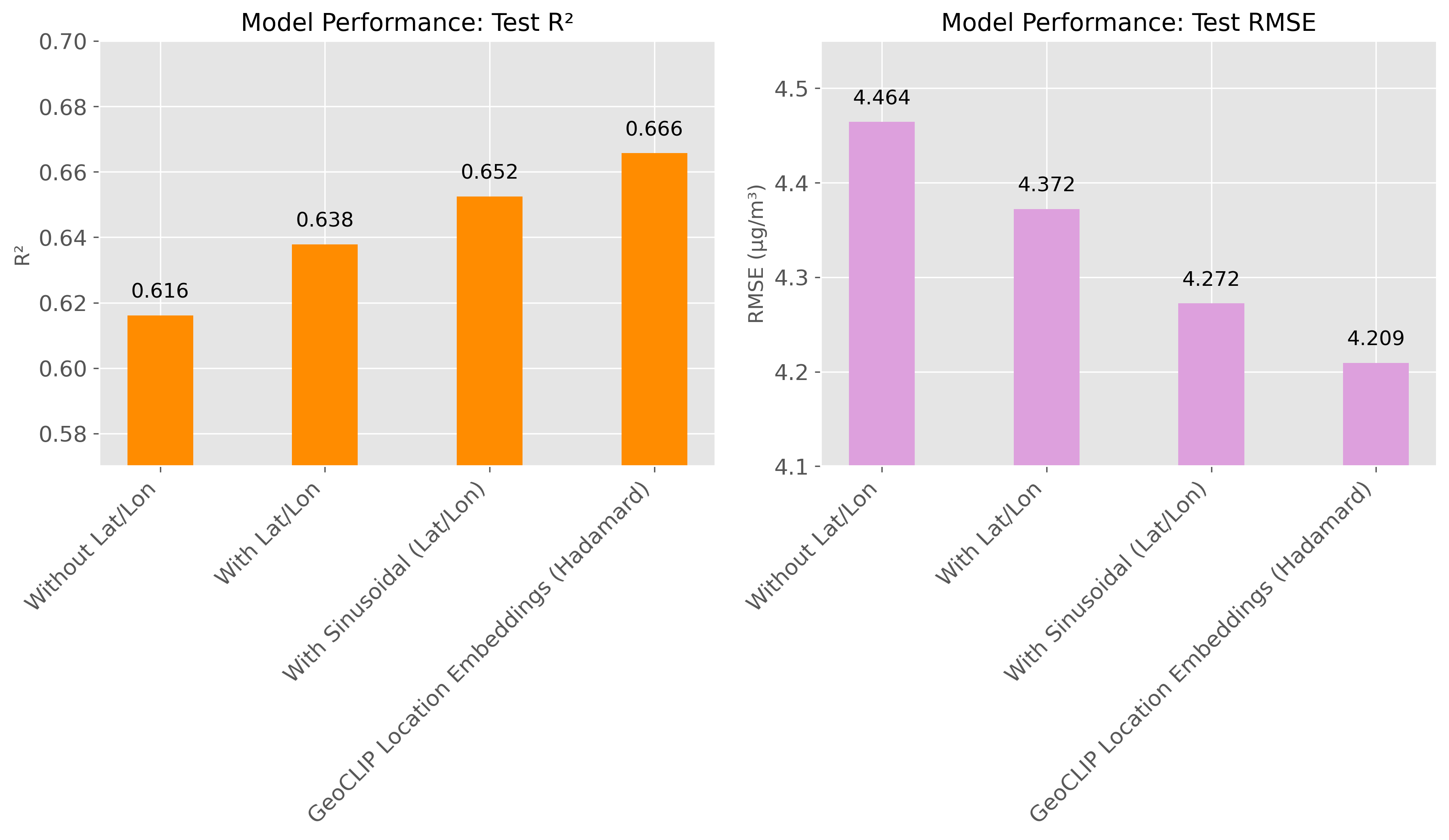}
    }

    \caption{Within-region model performance under Randomly (a) and Spatially (b) assigned test sets.}
    \label{fig:cv_metrics}
\end{figure}

Figure \ref{fig:cv_metrics} shows performance metrics for Within-Region scenarios. The results of the first three columns are not entirely surprising: adding geolocation information to the model while training enhances its ability to generalize to test locations that are close to or are within the training regions. Sinusoidal wrapping of geographic coordinates helps compared to a simple direct incorporation of degree-measurements of geographic coordinates, even though the region of the study is contained within the continental United States, where latitudes range from approximately $24^\circ \text{N}$ to $49^\circ \text{N}$ and longitudes range from $125^\circ \text{W}$ to $67^\circ \text{W}$ without discontinuities (such as those at poles or the International Date Line). 

What is more interesting is that GeoCLIP-extracted location embeddings help improve performance even further. It is worth remembering the context: our predictive features do include multi-variate observations of AOD measurements and other ancillary variables, and the GeoCLIP-extracted embeddings are static vectors for each location, derived partially from images on Flickr related to each location during pretraining. Although GeoCLIP embeddings are not based on satellite observations, they nevertheless help improve test performance on unseen locations for within region evaluation scenarios. 
The WR test results indicate that geolocation does indeed add interpolative value regardless of the selected approach to incorporate location: Both the na\"{\i}ve incorporation of geographic coordinates or through pre-learned location encoder embeddings, although interestingly, the prelearned embeddings increase predictive performance more than direct incorporation of geographic coordinates. 

\rev{Another noteworthy observation is that, under the random test scenario, GeoCLIP with Hadamard fusion yields the smallest bias (lowest $|\mathrm{MBE}|$), followed by the model without lat/lon (Table~\ref{tab:S1_random}). Models using direct coordinate features (raw lat/lon or their sinusoidal transforms) exhibit higher bias, although none show practically significant systematic over- or under-estimation. Under the spatial split, the model without lat/lon achieves the smallest bias, while the model with GeoCLIP embeddings fusion display a slightly larger average bias, but with notably smaller standard deviations across splits, indicating more stable performance across spatial partitions (Table~\ref{tab:S2_spatial}).}

\subsection{Out-of-Region (OoR) Evaluation}
Geographic generalizability is broadly defined as the ability of a model trained in one region to maintain its predictive performance when deployed in another region (Our of Region - OoR), which may have distinct geographic characteristics. Robust geographic generalizability is a sought-after quality for geospatial deep learning tasks, where models are often required to make predictions in areas without sufficient training data. The ability to maintain high performance across regions enhances the model’s applicability for large-scale geospatial analysis or environmental monitoring, where predictions must remain reliable despite geographic variability.

\begin{figure}[htbp]
    \centering
    \subfloat[Checkerboard partition with $\delta = 8.0^\circ$\label{fig:cb8}]{
        \includegraphics[width=0.48\linewidth]{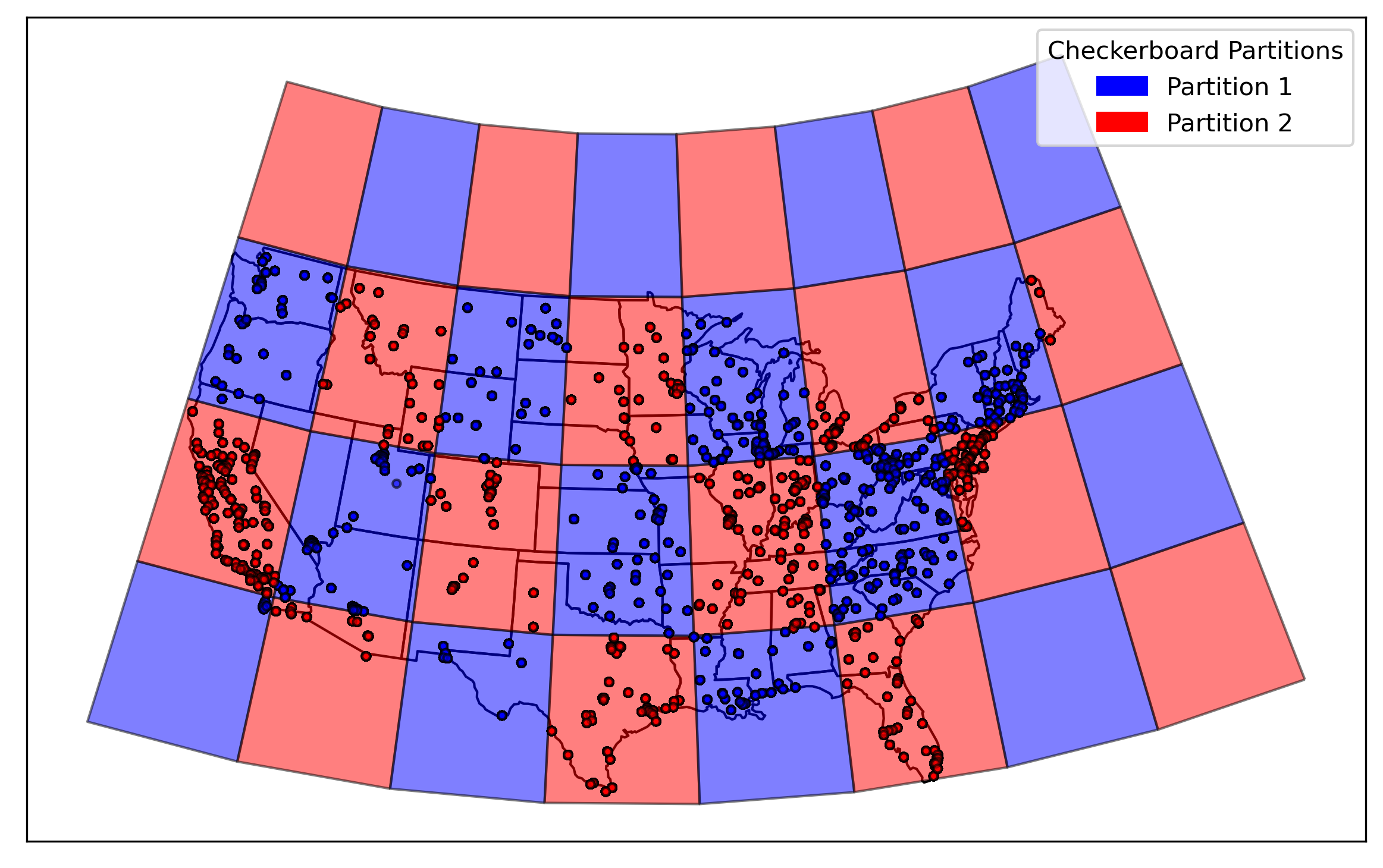}
    }
    \hfill
    \subfloat[Checkerboard partition with $\delta = 16.0^\circ$\label{fig:cb16}]{
        \includegraphics[width=0.48\linewidth]{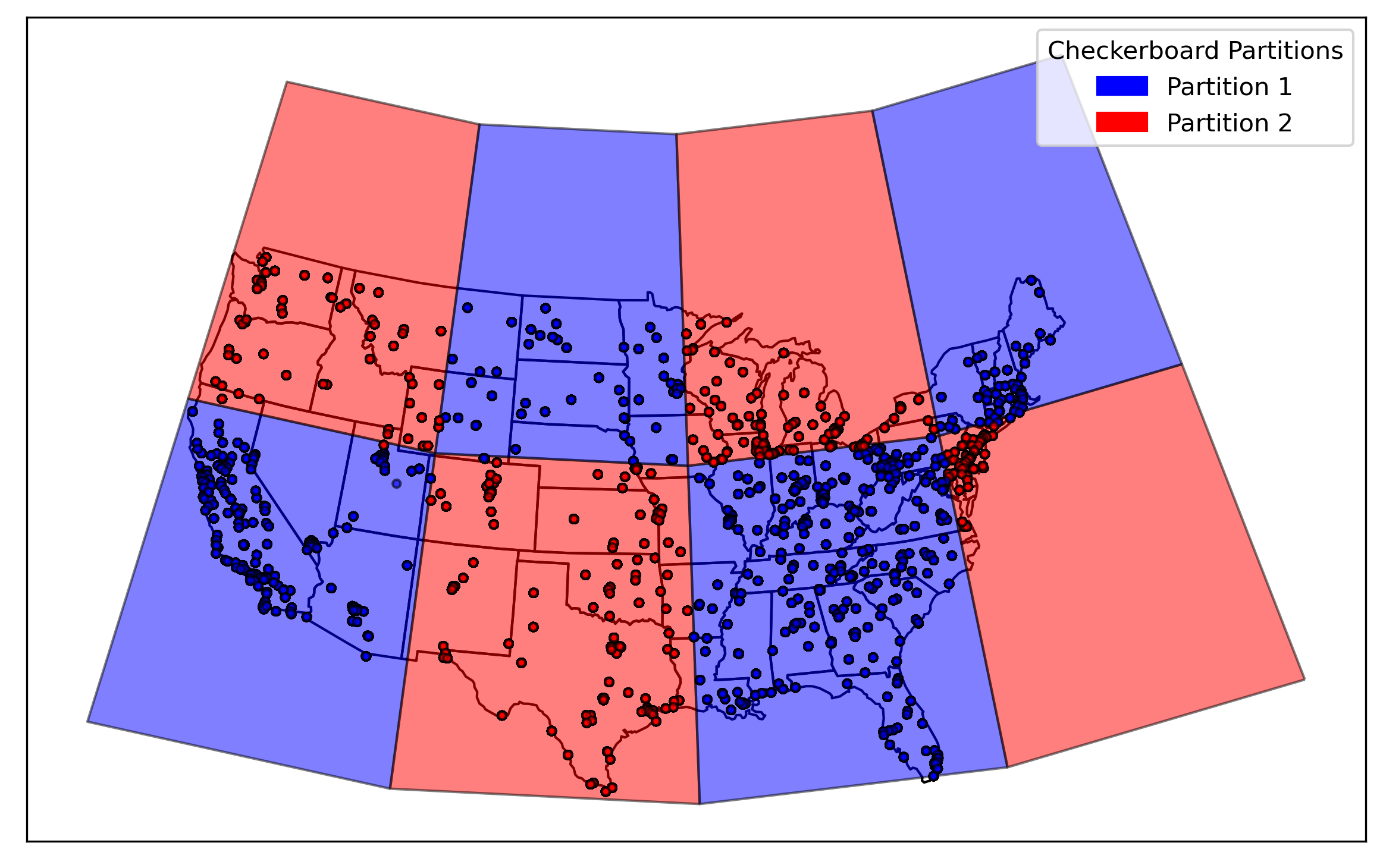}
    }
    \caption{Checkerboard spatial partitioning used for Out-of-region (OoR) evaluations. The circles show the location of AQS stations.}
    \label{fig:checkerboard}
\end{figure}

\textbf{Checkerboard Partition Evaluation}: For a rigorous OoR evaluation, we follow spatial partitioning designed by \cite{rolf2021generalizable} using a checkerboard pattern, training on one partition and testing on the other set (Figure \ref{fig:checkerboard}). This is to ensure that the training and testing sets are spatially non-overlapping and disjoint, allowing the test performance metrics to be a fair indication of model capacity to generalize to unseen regions.

Increasing the blocks' width (denoted by $\delta$, measured in degrees) increases the average distance between training and test observations, creating a more challenging test of the model's OoR geographic generalizability capabilities \cite{rolf2021generalizable}. Our experiments examine two checkerboard configurations with $\delta=8^{\circ}$ and $\delta=16^{\circ}$, corresponding to moderate and more extreme spatial validation, respectively (Figure \ref{fig:checkerboard}).

\begin{figure}[htbp]
    \centering
    \subfloat[Checkerboard test set metrics with $\delta = 8.0^\circ$\label{fig:checkerboard8}]{
        \includegraphics[width=0.85\textwidth]{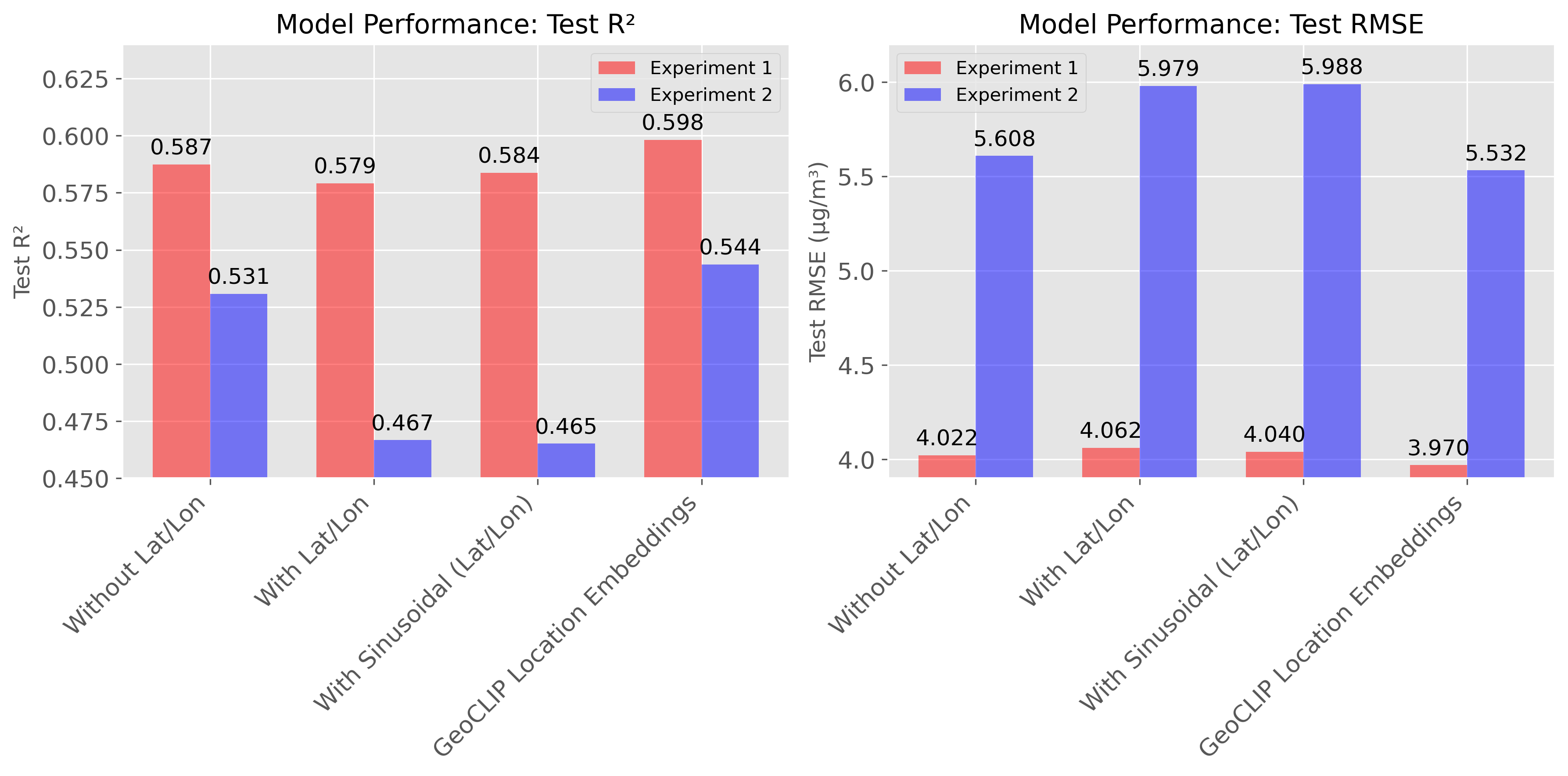}
    }

    \vspace{0.3cm}

    \subfloat[Checkerboard test set metrics with $\delta = 16.0^\circ$\label{fig:checkerboard16}]{
        \includegraphics[width=0.85\textwidth]{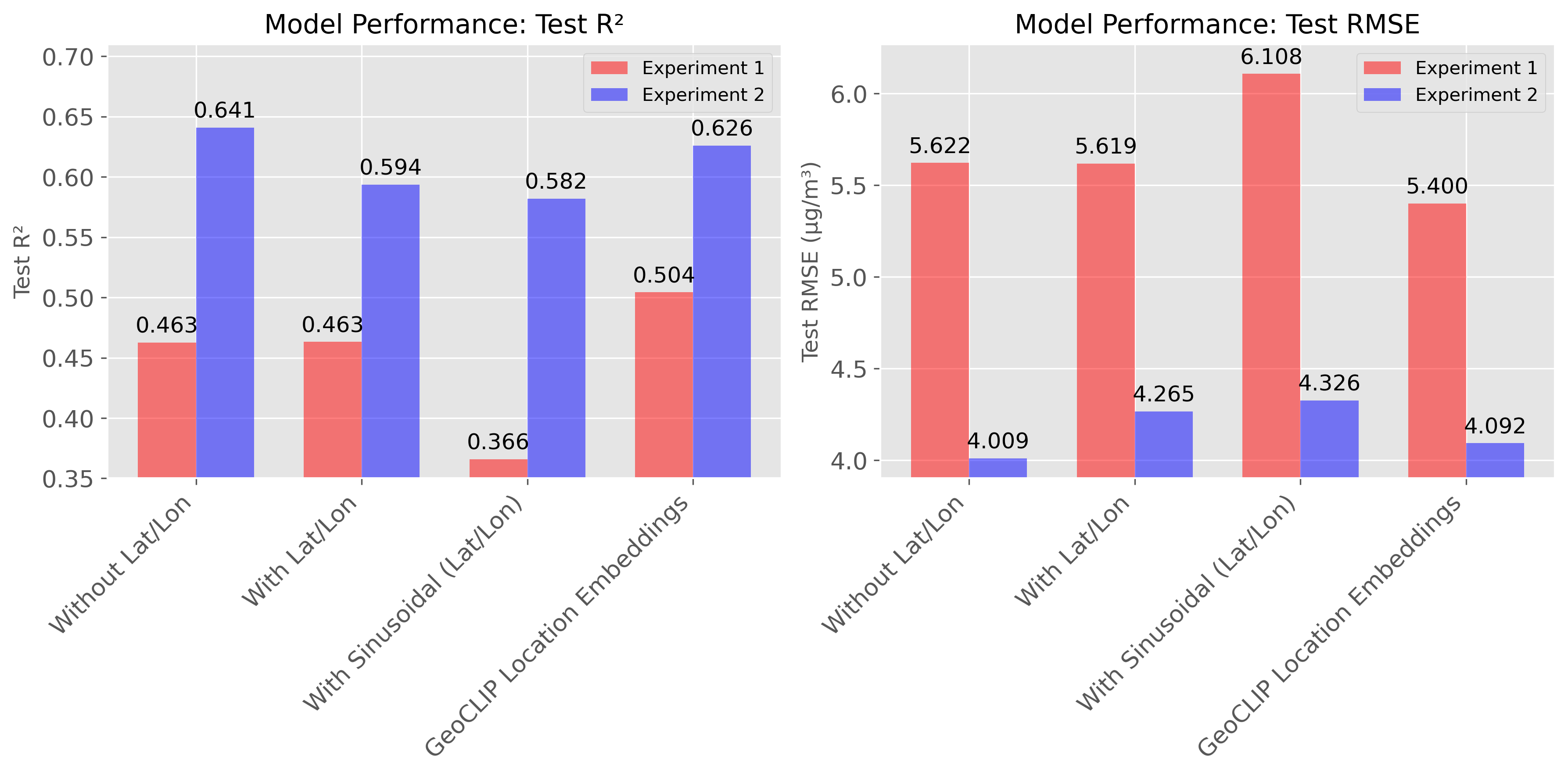}
    }

    \caption{Out-of-region evaluation using geospatial checkerboard partitioning with $\delta = 8.0^\circ$ (a) and $\delta = 16.0^\circ$ (b).}
    \label{fig:checkerboard_metrics}
\end{figure}

Figure \ref{fig:checkerboard_metrics} presents performance metrics under these settings, with the color of the bars corresponding to the \textit{test} partition colors on the maps. Put differently, for each value of $\delta$, we conduct two experiment sets (with five runs for each), one with the blue partition (1) as training set and red partition (2) as test set, and another experiment with swapping these sets. This is to control for geographic variation in the location of AQS stations, and the uneven number of stations in the partitions, which is inevitable in this real-world scenario. Please also note that the results here cannot be compared to the previous section, as partitioning leads to a significant drop in the number of available training stations. Nevertheless, comparisons within each bar chart are valid (but not across the bar charts). 

As can be seen in Figure \ref{fig:checkerboard_metrics}, the models without geolocation features and the models using GeoCLIP embeddings perform the best in OoR evaluations. The models without geolocation features learn the mapping of time-series observations to daily targets, avoiding  overfitting to specific regions, while GeoCLIP embeddings (in conjunction with time-series observations) appear to provide rich geographic information that allow for OoR generalization. It is worth remembering that GeoCLIP location encoders are pretrained, and even though the test regions in the checkerboard are not seen during the training of the \textit{downstream} air pollution estimation Bi-LSTM model, those regions are \textit{not} excluded when training GeoCLIP. Location encoders, after all, aim to distill information for every location into an embedding vector for downstream use. 

The considerable drop in performance for models using na\"{\i}ve geographic coordinates-with or without sinusoidal wrapping-suggests that the model has learned to associate specific locations with PM$_{2.5}$ levels, rather than relying on underlying aerosol or meteorological observations. Put differently, instead of using location to disambiguate when observations are rather similar, the model overfits to spatial patterns present in the training region—reducing its ability to generalize to unseen areas.

GeoCLIP embeddings, provide the best results under the $\delta = 8.0^\circ$ partition. However, for $\delta = 16.0^\circ$, which present a bigger challenge to geographic generalizability, models with no geolocation feature or embedding generalize the best to the test set, with GeoCLIP-enhanced models closely following in performance. For both $\delta$ values, models with lat/lon or sinusoidal wrappings of lat/lon see a drop in performance in OoR evaluation, which is contrary to the results seen in WR evaluations. These results highlights that na\"{\i}ve inclusion of geolocation features and insufficient evaluation schemes that do not span WR and OoR scenarios, can hinder geographic generalizability.

Results indicate that fusing embeddings from location encoders (in this case, GeoCLIP) improves generalization even across distant regions, although the benefits somewhat diminish as the partition size increases (from $\delta = 8.0^\circ$ to $\delta = 16.0^\circ$). Nevertheless, GeoCLIP embeddings lead to higher performance in three out of four checkerboard scenarios for such a temporally-dynamnic estimation scenario (with observations and targets that change day to day), reinforcing the value of pretrained location encoders for better generalization by distilling complex spatial attributes, even when training and test regions are spatially disjoint. The static embeddings derived from crowd-sourced imagery (e.g., in this case, Flickr) appear to encapsulate contextual information that improves predictions in geographically distant and non-overlapping regions by a model leveraging inputs of spatiotemporally-varying observations of aerosols.  

\rev{Across both checkerboard partitions, especially for $\delta{=}16^\circ$, GeoCLIP with Hadamard fusion maintains the smallest bias with similar values  between the two held-out folds (partitions), whereas direct coordinate features lead to larger and more asymmetric biases (Tables~\ref{tab:S3_cb8}–\ref{tab:S4_cb16}). This further shows that pretrained location embeddings enhance spatial generalization, consistent with other findings.}

\add{
To further assess the model’s geographic generalizability under a more extreme spatial configuration, we conducted an additional experiment using a checkerboard partition with $\delta = 30^\circ$. This partitioning effectively divides CONUS into western and eastern halves.
}
\add{
However, this configuration presents two key caveats. First, incorporating raw or sinusoidal geographic coordinates (e.g., latitude and longitude) in such a setting is inappropriate. Specifically, the longitude values in the test region become entirely out-of-distribution (OoD) relative to the training region, undermining the utility of direct coordinate-based features (our checkerboard evaluations earlier was partially motivated to prevent this from occurring). As a result, we do not train models with naive or sinusoidal lat/lon inputs in this evaluation. Second, PM$_{2.5}$ dynamics exhibit well-known and stark differences between eastern and western CONUS, most notably, due to differing emissions profiles and the prevalence of wildfire smoke in the West. Therefore, in addition to geographic coordinates being out of distribution, the target value follows different dynamics too. While understanding these caveats, we can still compare models with no geolocation features against the ones enhanced with fused GeoCLIP embeddings. 
}

\add{
For the Experiment 1 group (train on West, test on East), the model with GeoCLIP embeddings outperforms the model without lat/lon, achieving a higher average $R^2$ of 0.47 (vs. 0.38) and a lower average RMSE of 6.90 (vs. 7.43 $\mu$g/m$^3$). Similarly, in the Experiment 2 group (train on East, test on West), the GeoCLIP model again leads with an $R^2$ of 0.61 (vs. 0.52) and RMSE of 3.21 (vs. 3.56 $\mu$g/m$^3$). In both cases, the GeoCLIP-encoding-augmented models exhibit better generalization to the geographically disjoint test regions.
}
\add{
These results further demonstrate the effectiveness of using pretrained GeoCLIP embeddings in enhancing out-of-region performance, even under significant spatial disjointedness and differing aerosol dynamics between the western and eastern U.S. The embeddings likely encode semantic geographic context that aids the model in adapting to previously unseen regions. Despite the distribution shift and spatial disjointedness, GeoCLIP embeddings continue to provide semantically meaningful geographic context that boosts predictive performance across CONUS-wide OoR scenarios.
}

\subsection{Ablation}
\label{subsec:Ablation}
There are currently not  many pretrained location encoders available, as this is a burgeoning area of research. We considered the ones available, including SatCLIP \cite{klemmer2023satclip}, GeoCLIP  \cite{vivanco2023geoclip}, and CSP \cite{mai2023csp} . 
For our experiments we considered using location encoders of GeoCLIP and SatCLIP, given that in \cite{klemmer2023satclip} evaluations, CSP showed considerably lower comparative performance. However, it must be noted that CSP was one of the pioneer papers in this field.

GeoCLIP uses Fourier features for its positional encoder, has an embedding dimension of 512, and uses Flickr images to distill information into location embeddings, which consist of natural imagery taken by ordinary users. SatCLIP, on the other hand, use spherical harmonics, has an embedding space of dimension 256, and more importantly, uses Sentinel-2 satellite imagery to train the model, with a focus on global sampling of the images and locations. 

\begin{figure}[htbp]
    \centering
    \includegraphics[width=1\textwidth]{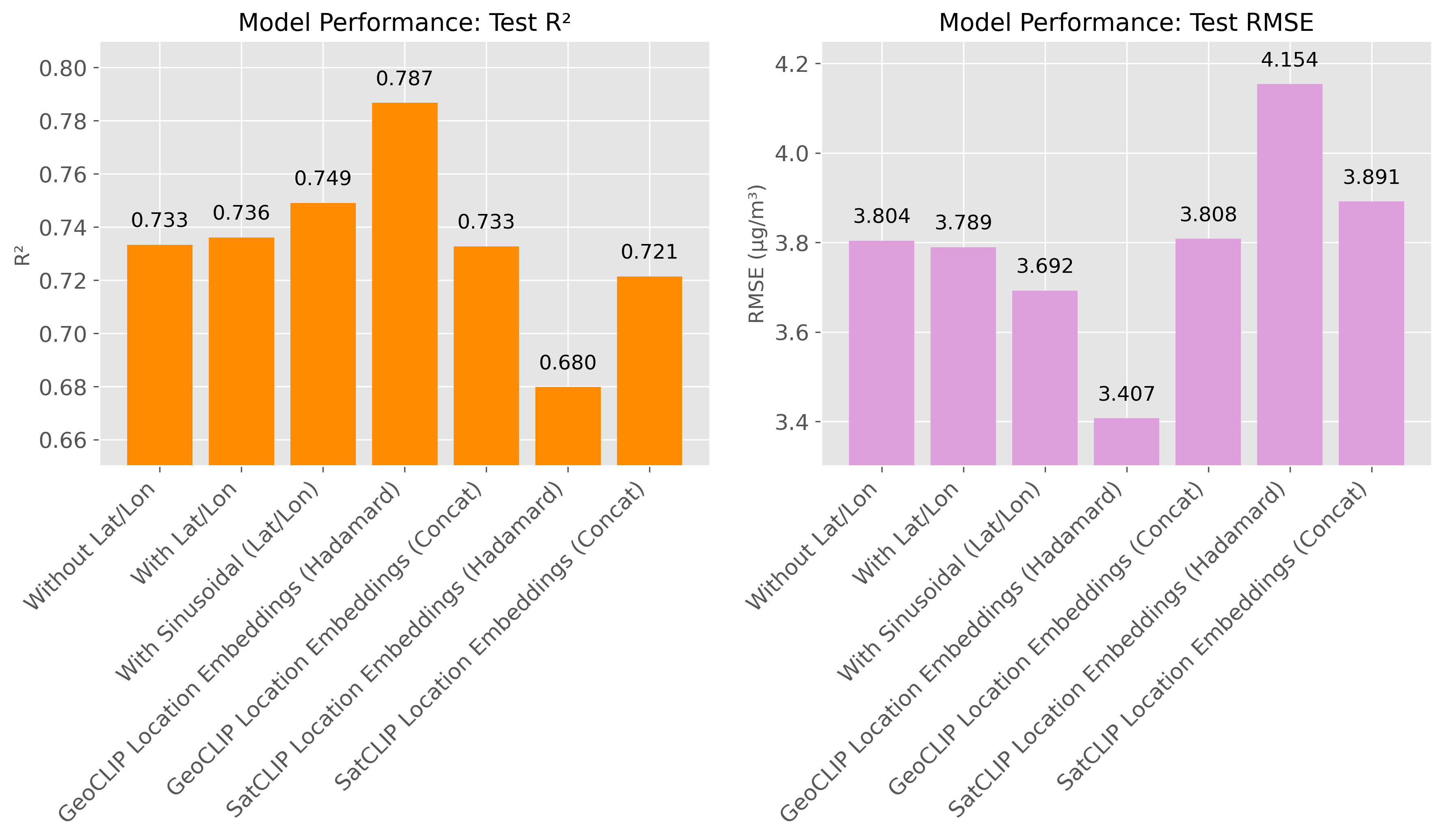}
    \caption{Ablation results on the random test set. Hadamard fusion strategy performs the best for location encoders, and GeoCLIP notably outperforms SatCLIP for our task.}
    \label{fig:ablation_random}
\end{figure}

There are two general strategies for fusing pre-learned location embeddings with other predictive features. The downstream model $f(\cdot)$ then learns to combine these embeddings with task-specific features for improved predictions, while benefiting from the spatial knowledge that was already encapsulated in $e_s$.
In our deep learning pipeline for estimating daily PM$_{2.5}$ concentrations, the location embedding $e_s$ is fused with the temporal feature representation $e_{\text{ts}}$ output by the Bi-LSTM. A typical fusion strategy involves either concatenation:
\[
y = f\left(e_{\text{ts}} \oplus e_s\right)
\]
or elementwise (Hadamard) product:
\[
y = f\left(e_{\text{ts}} \odot e_s\right)
\]
where $f(\cdot)$ denotes the final prediction head (e.g., a fully connected layer). Figure \ref{fig:ablation_random} shows the results of experiments with both GeoCLIP and SatCLIP using both fusion strategies.

While GeoCLIP shows superior performance compared to SatCLIP (Figure \ref{fig:ablation_random}), it is worth remembering that these are not production-level models yet, and rather, research products. For instance, the embedding dimension of 512 for GeoCLIP versus the 256 dimension for SatCLIP may impact their performance in our task, rather than the underlying positional encoders in each or training strategies. As more research is dedicated to location encoders in the near future, we expect more integration and evaluation in real-world tasks. As it relates to our task, however, it is worth discussing the underlying data of each location encoder. GeoCLIP \textit{may} outperform SatCLIP for PM$_{2.5}$ estimation tasks because its location embeddings are pretrained on Flickr imagery, which often captures ground-level human-centric scenes, including urban infrastructure, roads, and pollution-emitting facilities. These features are visually prominent and  related with PM$_{2.5}$ sources, allowing the model to encode relevant local context. In contrast, SatCLIP relies on Sentinel-2 imagery, which captures land cover from a top-down satellite perspective and \textit{may} miss finer-grained indicators of anthropogenic pollution sources (e.g., industrial stacks, traffic congestion), leading to less informative embeddings for this application.

\subsection{Qualitative Analysis of Spatial Patterns}
\label{subsec:qualitative analysis}
\begin{figure}[htbp]
    \centering
    \begin{minipage}{\textwidth}
        \centering
        \includegraphics[width=\linewidth, trim=54 13 10 10, clip]{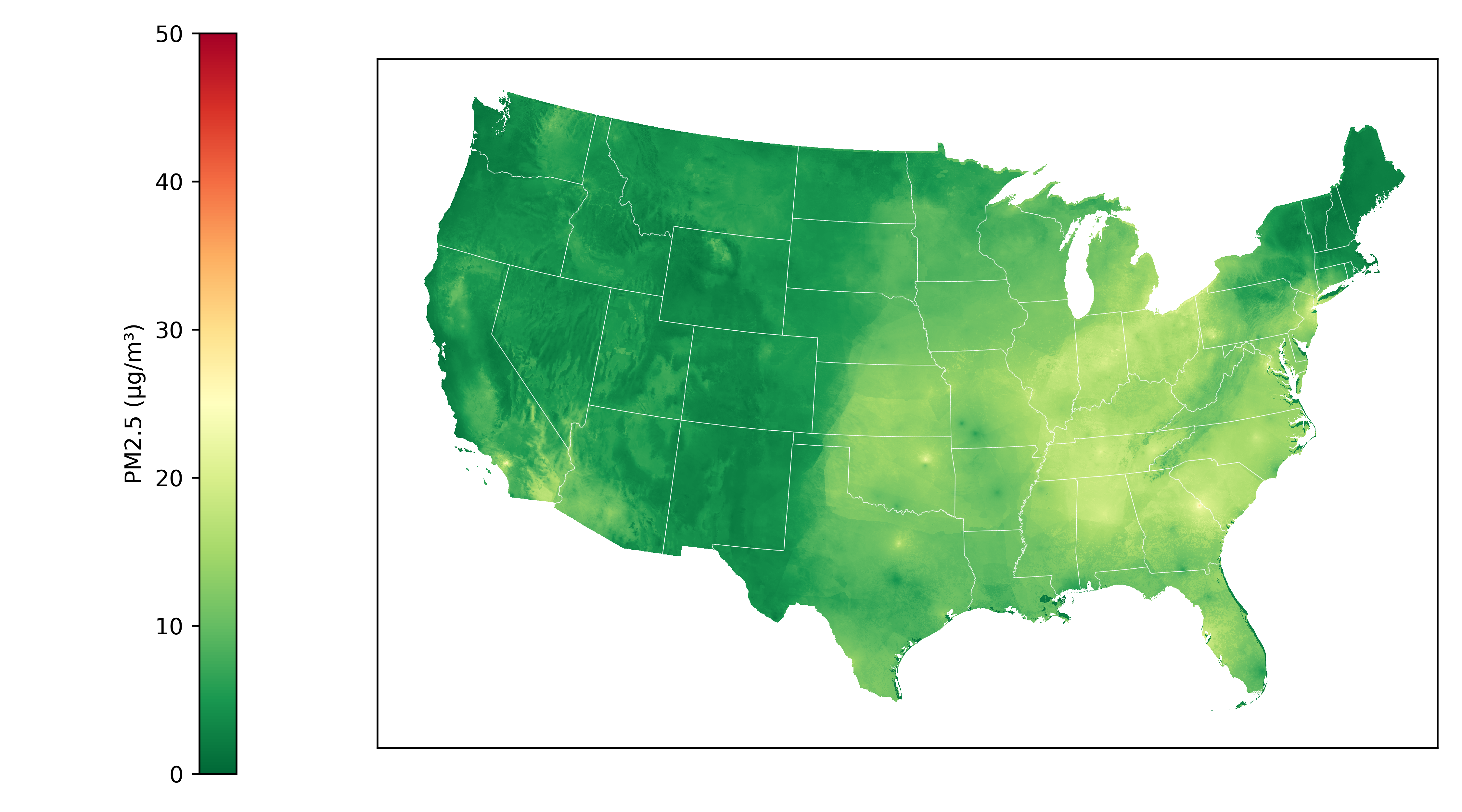}
        \vspace{2pt}
        (a) Without geolocation features
    \end{minipage}
    
    \vspace{0.1em}
    
    \begin{minipage}{\textwidth}
        \centering
        \includegraphics[width=\linewidth, trim=54 13 10 11, clip]{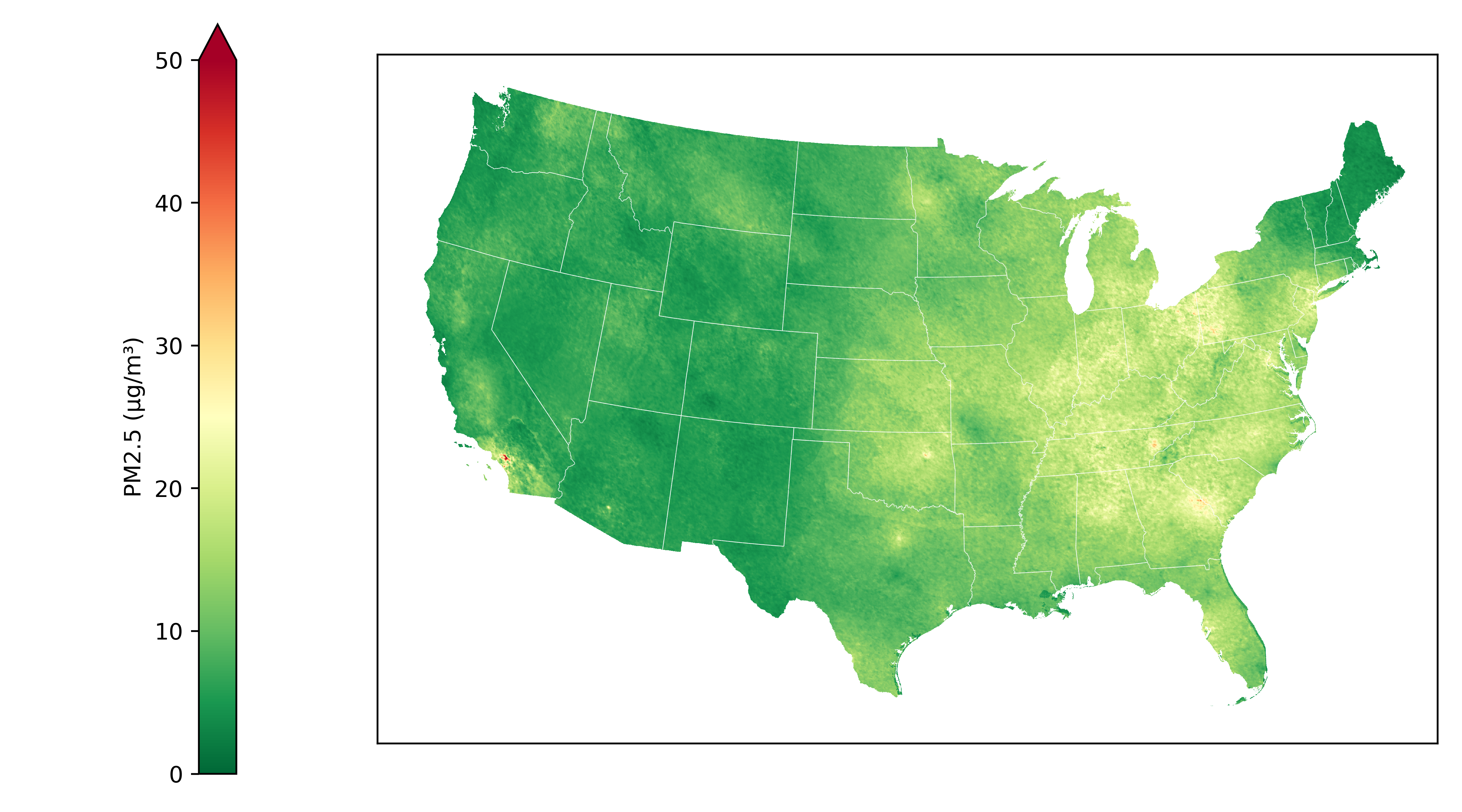}
        \vspace{2pt}
        (b) With pretrained location encoder (GeoCLIP)
    \end{minipage}
    
    \caption{Qualitative comparison of predicted PM$_{2.5}$ concentrations on July 5, 2021 across the contiguous U.S. Top (a): estimation using a model without geolocation features. Bottom (b): estimation using fused pretrained GeoCLIP location embeddings. Each map is generated at a spatial resolution of 1~km.}
    \label{fig:qualitative_july5}
\end{figure}

While the summary performance metrics provided in the previous sections shed light on the overall estimation performance of the models, we can visualize the model output over space. However, it is worth remembering that there is no 'ground truth' available for the entire space, i.e., there is no satellite can observe the surface-level PM$_{2.5}$ concentrations at all locations against which the model output can be compared. Figure~\ref{fig:qualitative_july5} provides a comparison of the maps generated using two different models, one without any geolocation feature (a), and (b) with pretrained GeoCLIP features, which performed best in the OOR and WR evaluation scenarios.

Visual inspection of predicted PM$_{2.5}$ concentration maps (Figure~\ref{fig:qualitative_july5}) reveals differences between models with and without geolocation features, and reminds of us of the spatial distribution of (a) pretraining imagery used for GeoCLIP pretraining, and (b) AQS stations used in training the downstream fused Bi-LSTM. However, these manifest in the outcome very differently.

\begin{figure*}[htbp]
    \centering
    \includegraphics[width=\textwidth]{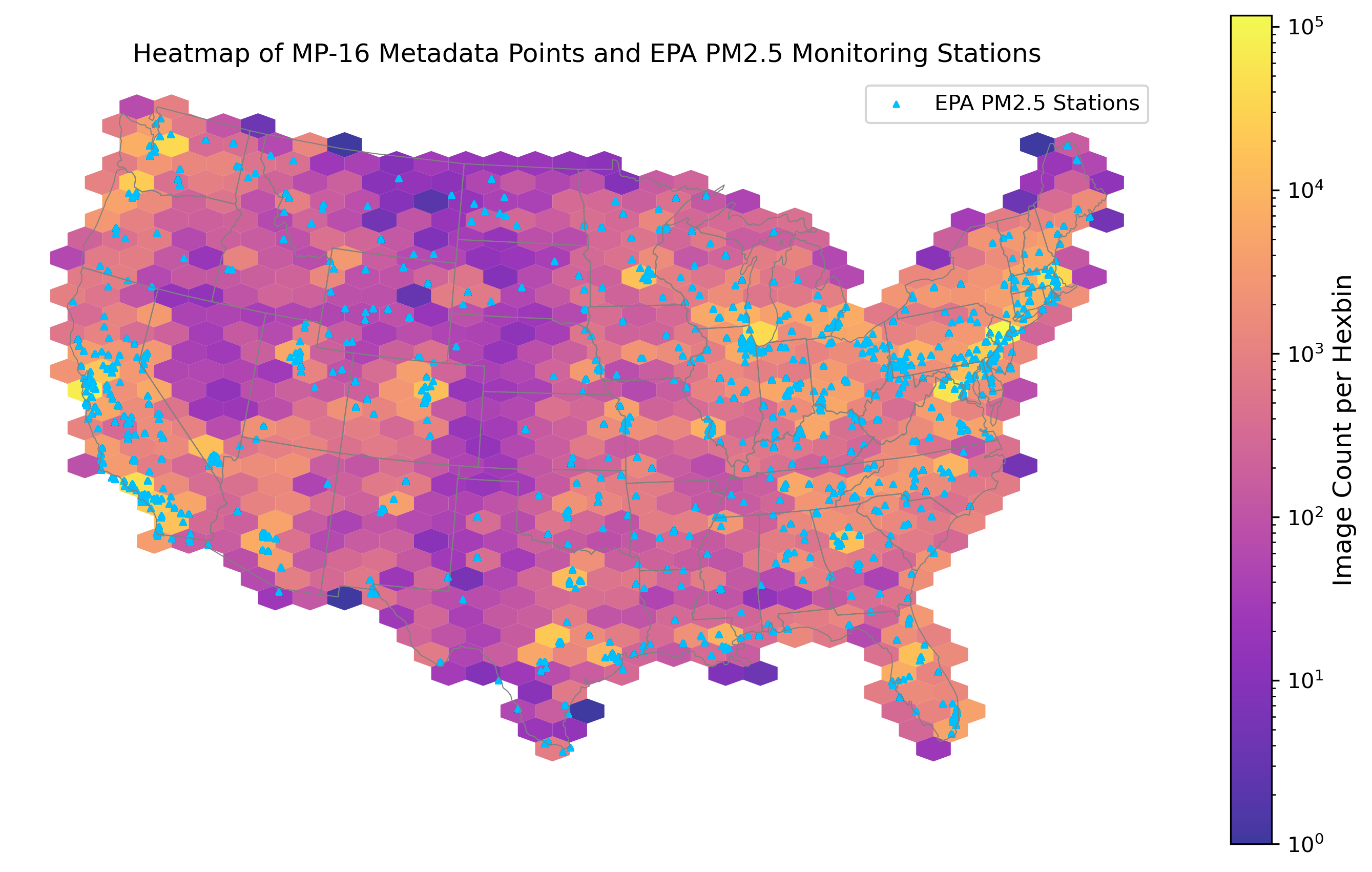}
    \caption{Spatial distribution of Flickr image density used in the MP-16 dataset (used for pretraining GeoCLIP \cite{vivanco2023geoclip}) overlaid with U.S. EPA PM$_{2.5}$ AQS monitoring stations. Flickr density is shown using a hexbin log-scale colormap; EPA monitoring stations are marked with cyan triangles. Both have higher densities in higher population areas.}
    \label{fig:mp16_epa_distribution}
\end{figure*}

The GeoCLIP-based model exhibits smoother and more spatially coherent patterns in several areas such the Great Plains States, where the model without geolocation tends to produce blocky artifacts with sharp an unnatural transitions, likely due to its inability to generalize across regions with sparse AQS measurement used in downstream Bi-LSTM training. This suggests that GeoCLIP’s pretrained location embeddings may help encode broad spatial context, enabling better geographic generalization. Interestingly, and of high importance to the air pollution estimation task (which is primarily intended to support health impact quantification in urban areas), in several urban centers—including Los Angeles, CA; Augusta, GA; Tulsa, OK; Pittsburgh, PA; and Cleveland, OH, all cities with known PM$_{2.5}$ pollution. The GeoCLIP-fused model captures elevated PM$_{2.5}$ levels at these known pollution hotspots, saturating the map color scale, rendering regions visibly red. The model without geolocation features seem to have smoother and lower value outputs in these places, even in Southern California. The fact that the model fused with GeoCLIP pre-learned embeddings is able to estimate higher values in urban areas counters a potential concern that spatial smoothing of static embeddings and reliance on static features would suppress (temporally dynamic) extremes, and highlights that the pretrained encoder can enhance urban prediction. This is in line with the quantitative results presented in the previous sections, where summary metrics pointed to the higher performance of GeoCLIP-fused models.  

On the other hand, the GeoCLIP output map seems to contain speckling or gaussian noise-like patterns in some rural or topographically complex areas. We believe this might be due to a combination of factors. First, the use of high-frequency Fourier features in the positional encoder can introduce localized instability, particularly when the downstream decoder $g(\phi(\varphi,\lambda))$ is applied to areas lacking strong pretraining samples. This is analogous to overfitting with high-degree basis functions, where unregularized regions may yield high-variance outputs. Put differently, the high-frequency positional encodings give the model the \textit{flexibility} to encode fine-scale spatial variation, but in regions lacking training signal, this flexibility can lead to high-variance or noisy predictions. Second, the underlying Flickr imagery used in GeoCLIP’s training is spatially uneven: denser in urban or tourist-heavy locations and sparser elsewhere. This uneven distribution may bias the learned representations and reduce embedding quality in under-sampled regions. Furthermore, fine-scale terrain-driven variations (e.g., along the Rockies) are more blurred in the GeoCLIP output. This may suggest that the model is not fully leveraging elevation or microclimate variation in those regions when fused with a static embedding. 

Figure~\ref{fig:mp16_epa_distribution} illustrates the spatial distribution of the Flickr images used in the MP-16 dataset—which are used in pretraining GeoCLIP embeddings—alongside the U.S. EPA PM$_{2.5}$ AQS monitoring network. It can generally be seen that the more populated an area is, the more AQS stations and the higher the density of Flickr pretraining images are (e.g., the Northeast, Southern California). The reason for higher density of Flickr images in populated areas is rather obvious: these are crowdsources, human-captured imagery. As for the spatial distribution of AQS stations, in the United States, PM$_{2.5}$ monitors operated by the U.S. Environmental Protection Agency (EPA) are used to assess compliance with the National Ambient Air Quality Standards (NAAQS)—federally mandated thresholds for ambient air pollutants intended to protect public health and the environment. Because most areas that are either currently designated as nonattainment or at risk of becoming nonattainment are urban and suburban, the majority of monitoring stations are concentrated in these regions. However, since the NAAQS also apply to Class I areas such as National Parks and Wilderness Areas, some monitors are sited in remote or protected landscapes to ensure compliance in those settings. Overall, sparsely monitored regions generally correspond to sparsely populated areas, including the Great Plains from North Dakota through West Texas, and the interior Southwest from Nevada through New Mexico.

This general alignment in spatial distributions partly explains the improvement in the overall estimation performance metrics as well as improved predictions of higher PM$_{2.5}$  concentration values in urban region. However, other regions such as the Mountain West exhibit Flickr image density sparsity (for pretraining) despite the presence of AQS monitors (for downstream task training). This spatial mismatch explains the limitation observed above, and highlights a limitation of pretrained location encoders: uneven coverage in the upstream pretraining dataset may result in lower-quality or biased embeddings in rural or topographically complex areas, potentially contributing to the spatial noise observed in those regions.

These qualitative analyses show the advantage and risks of location encoders pretrained on sparse data: they can enhance specificity and realism in familiar, well-sampled regions, but also amplify artifacts where spatial signal is weak or noisy, particularly when higher-frequency basis functions (such as Fourier features) are used in position encoding. Future research can perhaps focus on adaptive adjustment of positional encoder frequency degrees, particularly in lower-sampled regions.   

\subsection{Evaluation Under High-Concentration Conditions: The 2021 Dixie Fire}
\begin{figure*}[htbp]
\centering
\includegraphics[width=\textwidth]{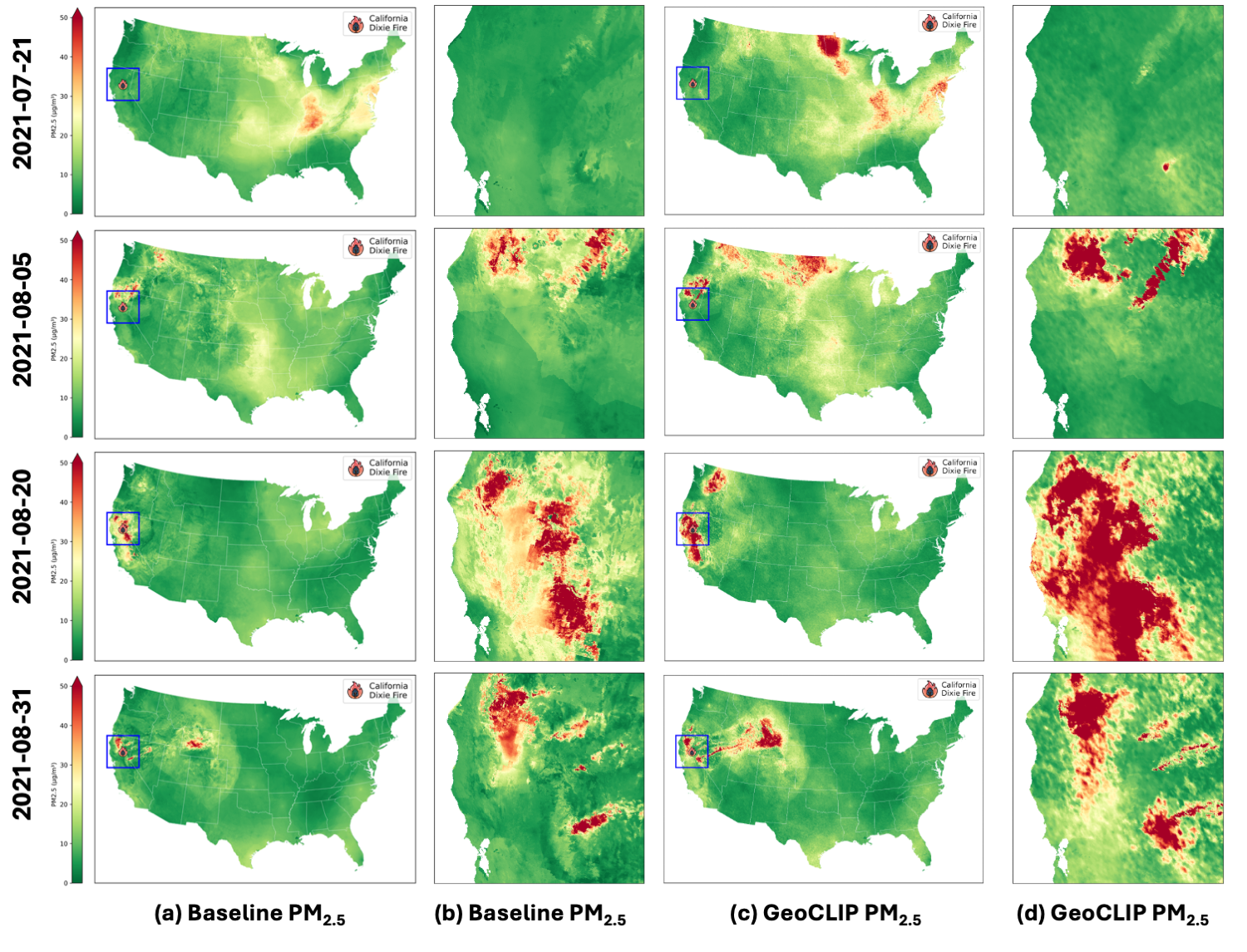}
\caption{\add{Estimated PM$_{2.5}$ during the 2021 Dixie Fire (Northern California) using baseline (no geographic features) and GeoCLIP-enhanced models. Each row shows a different date during peak wildfire activity. Columns (a) and (b) depict the baseline model output, while columns (c) and (d) show the GeoCLIP-enhanced model output at CONUS and regional scales. The GeoCLIP-enhanced model produces stronger plume intensity and spatial coherence over fire-affected regions \rev{and also captures elevated PM$_{2.5}$ concentrations over northern Minnesota on July~21, consistent with documented long-range smoke transport from concurrent western U.S. and Canadian wildfires.}}}

\label{fig:dixie_fire_preds}
\end{figure*}

The 2021 Dixie Fire was one of the largest and most destructive wildfires in California history, igniting on July 13, 2021 in the northern Sierra Nevada, near the Cresta Dam in Butte County. It scorched approximately 963,000 acres across five counties (Butte, Plumas, Tehama, Lassen, and Shasta). The fire destroyed nearly 1,300 structures and generated sustained high levels of PM$_{2.5}$ across large areas of the western United States. 
Figure~\ref{fig:dixie_fire_preds} compares the daily PM$_{2.5}$ estimations of the baseline model (without geolocation features) and the GeoCLIP-enhanced model over four dates spanning peak fire activity. Notably, the GeoCLIP-enhanced model produces larger, contiguous plumes with higher PM$_{2.5}$ concentrations over Northern California, better aligning with  fire-affected areas. In contrast, the baseline model exhibits more spatial fragmentation and muted intensity,  underestimating the scale and magnitude of pollution peaks. 
\rev{Interestingly, the GeoCLIP-enhanced model also captures elevated PM$_{2.5}$ concentrations over northern Minnesota after July 21, coinciding with the documented regional smoke event caused by wildfires in both western North America and Canada \citep{mn_pca_2021,axios_2021} at the time. These transported plumes led to air-quality alerts across the Upper Midwest, a pattern that the baseline model largely fails to reproduce.}

\rev{We acknowledge the persistence of potential artifacts, for instance, along NW to SE direction for the GeoCLIP-enhanced model, or in the case of  the fourth row of Figure~\ref{fig:dixie_fire_preds}, the relatively high values (light yellow) in disconnected patches surrounding the fire centers. In comparison, baseline model predictions look smoother across space, however, at the potential cost of underestimation at some high concentration areas. 
}

\begin{figure}[htbp]
    \centering
\includegraphics[width=\textwidth]{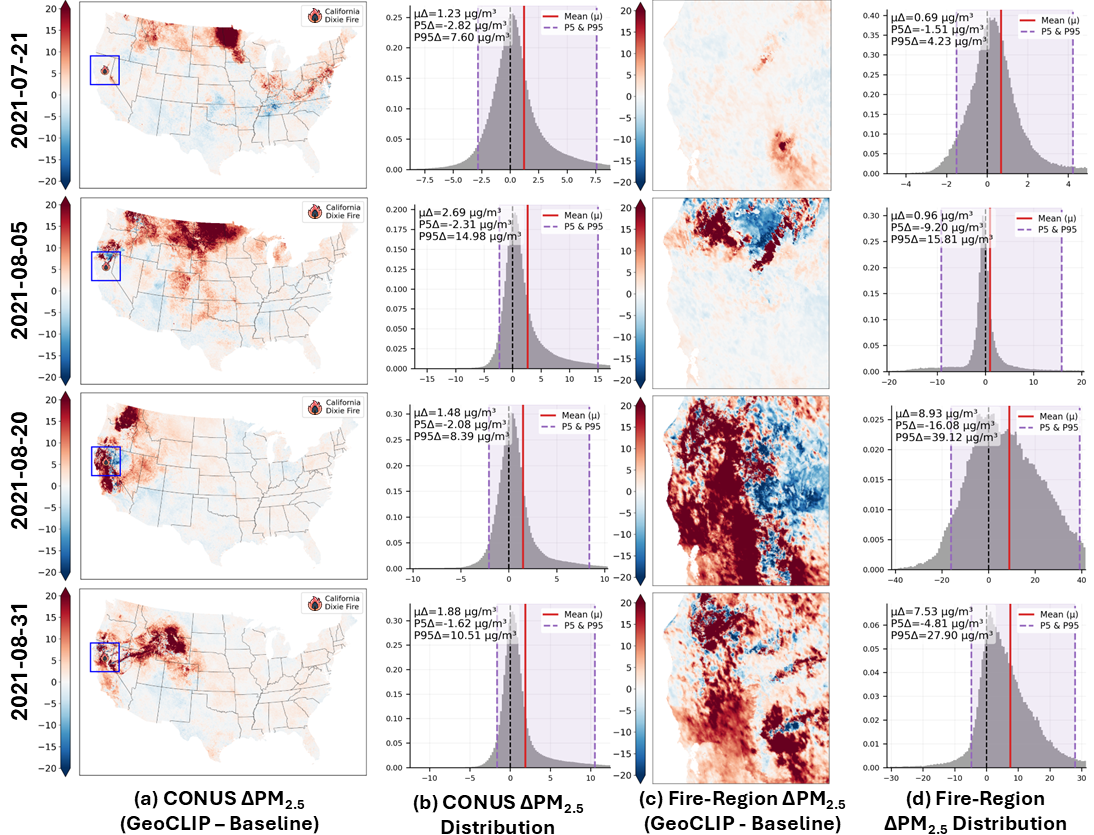}
    \caption{\rev{Differences in estimated PM$_{2.5}$ between GeoCLIP-enhanced and baseline models ($\Delta$PM$_{2.5}$ = GeoCLIP – Baseline) during the 2021 Dixie Fire. Columns (a) and (c) show spatial difference maps over CONUS and the fire-proximate region, respectively, while columns (b) and (d) display the corresponding pixelwise distributions. Positive (red) values indicate higher concentrations predicted by the GeoCLIP-enhanced model. Mean ($\mu_{\Delta}$) and percentile statistics (P5, P95) are reported for each date, demonstrating that GeoCLIP consistently produces higher PM$_{2.5}$ estimates over fire regions. }}
    \label{fig:dixie_fire_delta}
\end{figure}

\rev{To quantify the differences illustrated in Figure~\ref{fig:dixie_fire_preds}, Figure~\ref{fig:dixie_fire_delta} presents the pixelwise change in estimated PM${2.5}$ ($\Delta$PM${2.5}$ = GeoCLIP – Baseline) across CONUS and the fire region. On all four dates, the GeoCLIP-enhanced model estimates higher PM$_{2.5}$ values: Across CONUS, the mean difference is 1.23, 2.69, 1.48, and 1.88~µg/m$^3$ for July~21, August~5, August~20, and August~31, 2021, with P95 of 7.60, 14.98, 8.39, and 10.51~µg/m$^3$. Within the Northern California fire region, the mean differences are 0.69, 0.96, 8.93, and 7.53~µg/m$^3$ (showing small early differences and large late-episode gains), with P95 of 4.23, 15.81, 39.12, and 27.90~µg/m$^3$. These heavier and extended tails confirm that the GeoCLIP-enhanced model better captures the intensity and spatial extent of high-pollution plumes relative to the baseline model.}

\begin{figure*}[htbp]
\centering
\includegraphics[width=\textwidth]{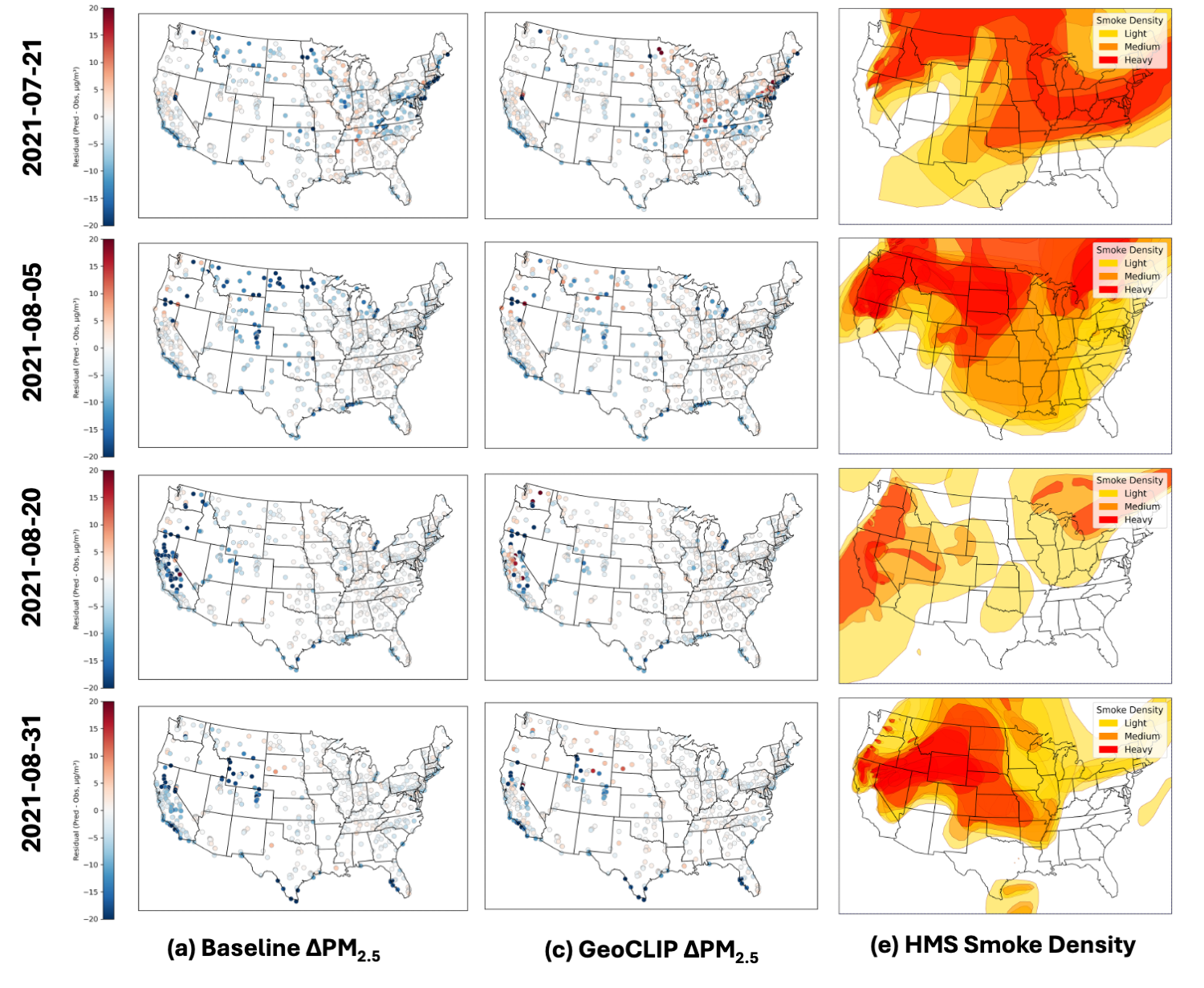}
\caption{\add{Residuals of PM$_{2.5}$ estimations (model estimate minus observed) during the 2021 Dixie Fire, with NOAA HMS smoke plumes overlaid. Baseline model which does not use geographic features (left) shows systematic underprediction (blue) in fire regions. GeoCLIP-enhanced model (middle) reduces negative residuals, and aligns better with the smoke plumes. Rightmost column shows HMS smoke density for visual reference.}}
\label{fig:dixie_fire_residuals}
\end{figure*}

\add{
Figure~\ref{fig:dixie_fire_residuals} further visualizes the spatial residuals for the same dates alongside NOAA HMS smoke density polygons. The GeoCLIP-enhanced model reduces systematic negative bias in smoke-affected regions, as seen in the transition from large blue residuals (underprediction) in the baseline model to more neutral residuals. This indicates an improved capacity to capture elevated pollution levels during dynamic wildfire events.
}
\add{
These observed patterns are consistent with the previous summary metrics and single-day maps: the GeoCLIP-fused model not only improves generalization across space, but also enhances larger value estimates under extreme air quality conditions. By encoding location semantics via pretrained embeddings, the model is better positioned to estimate higher concentrations in regions lacking dense observations but exhibiting complex pollution dynamics—such as those during wildfire plumes.
}

\add{
Despite improved performance, the GeoCLIP-enhanced model still underpredicts PM$_{2.5}$ in smoke-affected areas. This residual bias may stem from limited downstream training data in extreme pollution conditions, the static nature of GeoCLIP embeddings (which cannot reflect evolving fire behavior or atmospheric transport), and uneven upstream pretraining coverage. Fine-scale variability due to complex terrain or plume lofting (e.g., over the Sierra Nevada) may also be poorly captured by models relying on coarser inputs. These results suggest that while pretrained location embeddings improve spatiotemporal estimates, further gains may require integrating further contextual features into location encoders, as well as more strategies for enhancing extreme value estimations in downstream training.
}
%added

\section{Discussion}

\subsection{Estimation Performance Comparison}
We built on a previously validated Bi-LSTM with Attention model that achieves state-of-the-art performance in PM$_{2.5}$ estimation, including during high-concentration events \cite{wang2025high}. Compared to the Di et al.~\cite{di2019ensemble} dataset from 2005 to 2016, our model improves RMSE from 2.73 to 2.63 ($-$3.7\%) for all days, and 19.01 to 15.44 ($-$18.8\%) for high concentration days ($>$35~$\mu$g/m$^3$). Compared to Wei et al.~\cite{wei2023first} dataset which covers 2017 to 2021, for the same period, our base Bi-LSTM model improves the RMSE from 4.70 to 2.73 ($-$41.9\%) for all days, and 26.45 to 20.78 ($-$21.4\%) for high concentration days ($>$35~$\mu$g/m$^3$). Our Bi-LSTM with Attention explains 10\% more variance than our own Random Forest strong baseline, and 4\% more than a regular LSTM; RMSE is 3.59 (vs RF's 4.00, $-$10.3\%, and LSTM's 3.81, $-$5.8\%). More details and comparisons of the underlying air pollution estimation model can be found in \cite{wang2025high}. 

Using this strong model enabled us to focus specifically on the added value of geolocation features, without conflating gains from broader architectural or modeling differences. By building on an underlying model that already performs competitively against existing air pollution datasets, we strengthen the validity of our findings regarding the role of geolocation in enhancing generalizability and spatial transferability.

\subsection{On the Impact of Coordinate Transformation Methods}

In our experiments, for na\"{\i}ve incorporation of geolocation features (Approach 2), we applied sinusoidal wrapping to encode latitude and longitude. This approach preserves angular continuity while mapping coordinates into a bounded feature space. A natural question is whether alternative wrapping methods—such as those cataloged in \cite{mai2022review} are necessary. Our evaluation shows that the impact of coordinate wrapping on generalization and performance is largely invariant across such encodings (e.g., with raw coordinates vs. sinusoidal wrapping). The contiguous U.S. spans a continuous and bounded range of approximately $24^\circ$–$49^\circ$ N and $67^\circ$–$125^\circ$ W, without spatial discontinuities like poles or the International Date Line that would otherwise necessitate more complex encodings. Within this extent, all reasonable wrapping strategies—including sinusoidal transforms and Fourier projections—preserve relative spatial locality in similar ways. More importantly, our results show that the dominant factor influencing geographic generalizability is not the coordinate transformation itself, but whether coordinates are na\"{\i}vely passed to the model or replaced with pretrained higher-dimension embeddings such as those from GeoCLIP. Put differently, while the choice of encoding may affect inductive bias slightly, its influence is eclipsed by the broader modeling approach to incorporating geolocation.

Another question might be whether more expressive strategies like Fourier feature expansion \cite{tancik2020fourier}—would improve model performance, and more importantly, geographic generalizability.

The higher-frequency components of Fourier Expansion enable neural networks to learn high-frequency variations (i.e., higher spatial resolutions) in the input space by making the mapping more expressive, \add{capturing higher resolution variations.}

From a domain perspective, our target variable (daily PM$_{2.5}$ concentrations) and its geophysical covariates (e.g., meteorology, AOD, land cover) exhibit smooth spatial variation across the contiguous U.S. 
Additionally, since our model also incorporates temporal context and powerful sequence modeling (Bi-LSTM with Attention), the marginal impact of switching to a more expressive wrapping method on estimation performance is negligible. However, a more expressive wrapping method is sure to increase model capacity, which in turn, will inevitably hurt geographic generalizability, for similar reasons that we explain in the next paragraph. 

\subsection{Generalized Location Representations vs. Fine-tuning the Location Encoder}

Another related question is whether fine-tuning location encoders such as GeoCLIP on the downstream PM$_{2.5}$ estimation task would lead to performance gains. However, we intentionally avoided fine-tuning the location encoder to preserve and examine the generalization benefits that pretrained encoders are designed to provide. Fine-tuning poses two major risks in this context, defeating the purpose of using pretrained location encoders. First, the positional encoders that feed into location encoders (e.g., Fourier features with $k$ frequency bands) yield a much higher-dimensional representation than raw latitude and longitude (e.g., 16–64 features versus 2 or 4), substantially increasing model capacity and thus the risk of overfitting—especially when training data is sparse or regionally biased. This effect is exacerbated in our case where the downstream model must learn from limited and sparse PM$_{2.5}$ ground truth. For this same reason, a more expressive location wrapping method (such as Fourier features) will undoubtedly be more hurtful to geographic generalizability in the direct incorporation of location features (e.g., Approach 2) as well. 

Second, if the location encoder $g(\phi(\varphi,\lambda))$ is made trainable in the downstream task, it may begin to encode spurious location-specific patterns or implicitly compensate for missing covariates (e.g., unobserved emission sources or meteorological anomalies), which can entangle the learned representations with nuisance factors that do not generalize beyond the training spatial extent. This would undermine the very goal of location encoders: to produce generalized, reusable, and disentangled geographic representations distilled from large, unlabeled datasets during pretraining. Freezing the encoder ensures that the spatial attributes extracted from pretraining (e.g., Flickr imagery) remain invariant and generalize to new tasks. As our results confirm, using fixed embeddings from GeoCLIP improves both within-region and out-of-region performance—highlighting that the benefits of pretrained location encoders can stem from their consistency, not their adaptability to downstream noise or task peculiarities.

\section{Conclusion}
In this paper, we quantified the impacts of incorporating geolocation information in deep learning for temporally-dynamic estimation, with emphasis on geographic generalizability. We conducted a series of experiments by modifying a model for estimating daily average concentrations of PM$_{2.5}$ in the continental United States.  
We measured and compared within-region (WR) and Out-of-region (OoR) performance by incorporating geographic coordinates as features, as well as incorporating location embeddings from a more advanced GeoCLIP location encoder \cite{vivanco2023geoclip}. Our WR evaluations intended to examine whether incorporating geolocation information into deep learning enhances its interpolative ability (where test locations are roughly within the bounds of training locations). Our results indicated that adding geographic coordinates as features indeed appear to add interpolative value to the estimation task, while embeddings from the GeoCLIP location encoder even further enhanced WR evaluation performance. 

We also conducted experiments to evaluate OoR performance, by partitioning the geographic area into non-overlapping, disjoint areas that separate training and testing data. Our results highlight that  na\"{\i}ve inclusion of geographic coordinates as features can hinder model performance in OoR scenarios; however, well-designed location encoders such as GeoCLIP provide improvements in geographic generalizability. These findings show how geolocation features influence model behavior. In WR scenarios, geographic coordinates can serve as useful disambiguators—helping the model refine predictions when observations are spatially proximate. However, in OoR scenarios, these same features encourage overfitting by allowing the model to learn location-target associations rather than generalizable mappings from observations to target. This contrast highlights the value of incorporating location in ways that prioritize observational relevance rather than region-specific memorization.

In addition to quantitative results, we provided an in-depth discussion of the methods of incorporating geolocation into deep learning, along with analyses of variations and ablations, and the expected impacts on geographic generalizability. Improved geographic generalizability in PM$_{2.5}$ estimation has direct importance for public health and environmental policy. Models that can accurately estimate pollution levels across both data-rich and data-sparse regions support more equitable exposure assessments, especially in under-monitored or rural communities. This enhances the ability of policymakers to implement data-driven air quality regulations, evaluate compliance with federal standards, and better allocate resources for monitoring and mitigation efforts.

While the original intention of our work was to quantify the impacts of incorporating geolocation information into deep learning for the estimation of a target with day-to-day changes with multi-variate observations, our results also highlight the value of location encoders (despite and in addition to the multi-variate domain observations), and the value of an emerging research body to distill information about the Earth in latent embeddings for seamless incorporation into deep learning. This area of research is relatively young, and so far, has been focused on fusing one mode of data with location representations. To  distill even more information into these models, future research can move towards multi-modal and multi-sensor enhancements that better capture information about a place into these latent embeddings. As evidenced by our dynamic application that requires daily observation of aerosol content in the atmosphere, adding a temporal aspect to location encoders can also be invaluable in applications on dynamic phenomena.  

Further, our qualitative analysis revealed the impact of spatially-uneven pretraining samples and spatially-sparse downstream supervision values, resulting in noise artifacts or smoothed-out estimates in certain regions. The uneven distribution of pretraining samples combined with high-degree basis functions used in positional encoders of location encoders appears to result in speckle-like noise. Therefore, spatial regularization of positional encoders with adaptive scales promises to be a productive area for future research.

\section*{Data Availability Statement}
All data uses are openly available at \cite{karimzadeh2025data} \add{and all software code are available at \cite{karimzadeh2025code}}.

\section*{Funding}
The National Science Foundation grant number 2026962 and NIH/NIEHS grant R21ES032973 have supported this work. The content is solely the responsibility of the authors and does not necessarily represent the official views of the University of Colorado, NSF, NIH, or NIEHS.

\section*{Author Contributions}

\noindent
\textbf{Morteza Karimzadeh}: Conceptualization, Methodology, Formal analysis, Writing – original draft, Supervision. \\
\textbf{Zhongying Wang}: Data curation, Software, Validation, Methodology, Writing – review \& editing. \\
\textbf{James L. Crooks}: Resources, Supervision, Methodology, Writing – review \& editing.

\section*{Disclosure statement}
Authors have no conflict of interest to report.

\bibliographystyle{tfnlm}   
\bibliography{interactnlmsample}

\bigskip

\appendix

\section{Evaluation Metrics Tables}

% ---------- Table S1: Within-Region (Random split) ----------
\begin{table}[h]
\caption{\rev{Within-Region evaluation metrics for five random splits, each with a different training and testing subset. Mean~$\pm$~standard deviation are reported. Variation in metrics over the five splits reflects differences in subset difficulty and target distributions.}}
\label{tab:S1_random}
\centering
\footnotesize
\setlength{\tabcolsep}{4pt}
\renewcommand{\arraystretch}{1.15}
\begin{tabularx}{\textwidth}{l C{2.2cm} C{2.9cm} C{2.4cm}}
\toprule
\textbf{Model variant} & \textbf{Test $R^2$} & \textbf{Test RMSE ($\mu$g\,m$^{-3}$)} & \textbf{Test MBE ($\mu$g\,m$^{-3}$)}\\
\midrule
Without Lat/Lon & \(0.73\pm0.035\) & \(3.80\pm0.24\) & \(-0.06\pm0.72\) \\
With Lat/Lon & \(0.74\pm0.016\) & \(3.79\pm0.11\) & \(0.07\pm0.79\) \\
With Sinusoidal (Lat/Lon) & \(0.75\pm0.024\) & \(3.69\pm0.17\) & \(0.11\pm0.70\) \\
GeoCLIP Location Embeddings (Hadamard) & \(0.79\pm0.010\) & \(3.41\pm0.08\) & \(-0.03\pm0.32\) \\
GeoCLIP Location Embeddings (Concat) & \(0.73\pm0.036\) & \(3.81\pm0.25\) & \(0.26\pm0.93\) \\
SatCLIP Location Embeddings (Hadamard) & \(0.68\pm0.077\) & \(4.15\pm0.45\) & \(0.39\pm1.31\) \\
SatCLIP Location Embeddings (Concat) & \(0.72\pm0.024\) & \(3.89\pm0.16\) & \(-0.17\pm0.73\) \\
\bottomrule
\end{tabularx}
\end{table}

% ---------- Table S2: Within-Region (Spatial split) ----------
\begin{table}[h]
\caption{\rev{Within-Region evaluation metrics for five different spatial splits (with entire time series of test stations held out), each with a distinct training and testing subset. Mean~$\pm$~standard deviation are reported. Variation across splits reflects differences in spatial coverage and target distributions.}}
\label{tab:S2_spatial}
\centering
\footnotesize
\setlength{\tabcolsep}{4pt}
\renewcommand{\arraystretch}{1.15}
\begin{tabularx}{0.89\textwidth}{l C{2.2cm} C{2.9cm} C{2.4cm}}
\toprule
\textbf{Model variant} & \textbf{Test $R^2$} & \textbf{Test RMSE ($\mu$g\,m$^{-3}$)} & \textbf{Test MBE ($\mu$g\,m$^{-3}$)}\\
\midrule
Without Lat/Lon & \(0.62\pm0.13\) & \(4.46\pm0.81\) & \(0.05\pm1.63\) \\
With Lat/Lon & \(0.64\pm0.067\) & \(4.37\pm0.71\) & \(-0.16\pm1.04\) \\
With Sinusoidal (Lat/Lon) & \(0.65\pm0.074\) & \(4.27\pm0.64\) & \(-0.36\pm0.87\) \\
GeoCLIP Location Embeddings & \(0.67\pm0.049\) & \(4.21\pm0.65\) & \(0.28\pm0.47\) \\
\bottomrule
\end{tabularx}
\end{table}

% ---------- Table S3: Checkerboard (δ = 8°) ----------
\begin{table}[h]
\caption{\rev{Out-of-Region (Checkerboard, $\delta{=}8^\circ$): metrics on the two disjoint test partitions~1 and~2. 
The partitions contain different proportions of test stations—48.59\% in partition~1 and 51.41\% in partition~2—reflecting the inherently uneven geographic distribution of AQS monitoring sites across the contiguous United States.}}
\label{tab:S3_cb8}
\centering
\footnotesize
\setlength{\tabcolsep}{4pt}
\renewcommand{\arraystretch}{1.15}
\begin{tabularx}{\textwidth}{l C{2.8cm} C{3.4cm} C{3.2cm}}
\toprule
\textbf{Model variant} & \textbf{Test $R^2$ (1\textbar{}2)} & \textbf{Test RMSE ($\mu$g\,m$^{-3}$) (1\textbar{}2)} & \textbf{Test MBE ($\mu$g\,m$^{-3}$) (1\textbar{}2)} \\
\midrule
Without Lat/Lon             & 0.59\enspace\textbar\enspace0.53 & 4.02\enspace\textbar\enspace5.61 & -1.00\enspace\textbar\enspace0.27 \\
With Lat/Lon                & 0.58\enspace\textbar\enspace0.47 & 4.06\enspace\textbar\enspace5.98 & -0.51\enspace\textbar\enspace-1.42 \\
With Sinusoidal (Lat/Lon)   & 0.58\enspace\textbar\enspace0.47 & 4.04\enspace\textbar\enspace5.99 & 0.42\enspace\textbar\enspace-0.65 \\
GeoCLIP Location Embeddings & 0.60\enspace\textbar\enspace0.54 & 3.97\enspace\textbar\enspace5.53 & 0.19\enspace\textbar\enspace0.19 \\
\bottomrule
\end{tabularx}
\end{table}

% ---------- Table S4: Checkerboard (δ = 16°) ----------
\begin{table}[h]
\caption{\rev{Out-of-Region (Checkerboard, $\delta{=}16^\circ$): metrics on the two disjoint test partitions~1 and~2. 
The partitions contain different proportions of test stations—63.51\% in partition~1 and 36.49\% in partition~2—reflecting the inherently uneven geographic distribution of AQS monitoring sites across the contiguous United States.}}
\label{tab:S4_cb16}
\centering
\footnotesize
\setlength{\tabcolsep}{4pt}
\renewcommand{\arraystretch}{1.15}
\begin{tabularx}{\textwidth}{l C{2.8cm} C{3.4cm} C{3.2cm}}
\toprule
\textbf{Model variant} & \textbf{Test $R^2$ (1\textbar{}2)} & \textbf{Test RMSE ($\mu$g\,m$^{-3}$) (1\textbar{}2)} & \textbf{Test MBE ($\mu$g\,m$^{-3}$) (1\textbar{}2)} \\
\midrule
Without Lat/Lon             & 0.46\enspace\textbar\enspace0.64 & 5.62\enspace\textbar\enspace4.01 & -1.68\enspace\textbar\enspace-0.15 \\
With Lat/Lon                & 0.46\enspace\textbar\enspace0.59 & 5.62\enspace\textbar\enspace4.26 & -0.34\enspace\textbar\enspace-1.14 \\
With Sinusoidal (Lat/Lon)   & 0.37\enspace\textbar\enspace0.58 & 6.11\enspace\textbar\enspace4.33 & -2.19\enspace\textbar\enspace-0.87 \\
GeoCLIP Location Embeddings & 0.50\enspace\textbar\enspace0.63 & 5.40\enspace\textbar\enspace4.09 & -0.03\enspace\textbar\enspace0.26 \\
\bottomrule
\end{tabularx}
\end{table}

\end{document}